\documentclass[runningheads]{llncs}

\usepackage[mobile]{accv}

\usepackage{accvabbrv}
\usepackage{wrapfig}

\usepackage{graphicx}
\usepackage{booktabs}
\usepackage{dcolumn} 

\usepackage[dvipsnames]{xcolor}
\usepackage{amsmath} %
\usepackage[T1]{fontenc}
\usepackage[utf8]{inputenc}
\usepackage{algorithm}
\usepackage{algpseudocode}

\usepackage{listings}
\usepackage{xcolor} 

\usepackage{booktabs}
\usepackage{array}
\usepackage{subcaption} %

\newcommand{\comment}[1]{}

\DeclareUnicodeCharacter{0229}{\c{e}}

\newcommand{\Var}[1]{\text{Var}(#1)}
\newcommand{\E}[1]{\operatorname{E}}

\definecolor{mycolor}{HTML}{2F6BAA}

\usepackage[accsupp]{axessibility}  

\usepackage{hyperref}

\usepackage{orcidlink}

\begin{document}

\title{D'OH: Decoder-Only Random Hypernetworks \\ for Implicit Neural Representations}

\titlerunning{Decoder-Only Hypernetworks (D'OH)}

\author{Cameron Gordon\inst{1}  \and
Lachlan E. MacDonald\inst{2} \and
Hemanth Saratchandran\inst{1} \and 
Simon Lucey\inst{1}}

\authorrunning{C. Gordon et al.}

\institute{Australian Institute for Machine Learning,\\ University of Adelaide, Adelaide SA 5000, Australia\\ 
\email{\{first.last\}@adelaide.edu.au}\\ \and
Mathematical Institute for Data Science,\\ John Hopkins University, Baltimore MD 21218, USA
}

\maketitle

\begin{abstract}
  Deep implicit functions have been found to be an effective tool for efficiently encoding all manner of natural signals. Their attractiveness stems from their ability to compactly represent signals with little to no offline training data. Instead, they leverage the implicit bias of deep networks to decouple hidden redundancies within the signal. In this paper, we explore the hypothesis that additional compression can be achieved by leveraging redundancies that exist \emph{between} layers. We propose to use a novel runtime decoder-only hypernetwork -- that uses no offline training data -- to better exploit cross-layer parameter redundancy. Previous applications of hypernetworks with deep implicit functions have employed feed-forward encoder/decoder frameworks that rely on large offline datasets that do not generalize beyond the signals they were trained on. We instead present a strategy for the optimization of runtime deep implicit functions for single-instance signals through a Decoder-Only randomly projected Hypernetwork (D'OH). By directly changing the latent code dimension, we provide a natural way to vary the memory footprint of neural representations without the costly need for neural architecture search on a space of alternative low-rate structures. 
  
\end{abstract}

\section{Introduction}
\label{sec:intro}

Implicit neural representations (INRs), also known as coordinate networks, have received attention for their ability to represent signals from different domains -- including sound, images, video, signed distance fields, and neural radiance fields -- within a signal-agnostic framework \cite{sitzmann_implicit_2020,xie_neural_2022,mescheder_occupancy_2019, dupont_coin_2021,mildenhall_nerf_2020,chen_nerv_2021}. When combined with quantization strategies, they act as a neural signal compressor which can be applied generally across different modalities -- the best example of which is Compressed Implicit Neural Representations (COIN)\cite{dupont_coin_2021}. Combining this with lossless entropy compression can lead to even further reductions \cite{strumpler_implicit_2022, dupont_coin_2021, schwarz_modality-agnostic_2023,gordon_quantizing_2023, damodaran_rqat-inr_2023}.

\begin{figure}
    \centering
    \includegraphics[width=\linewidth]{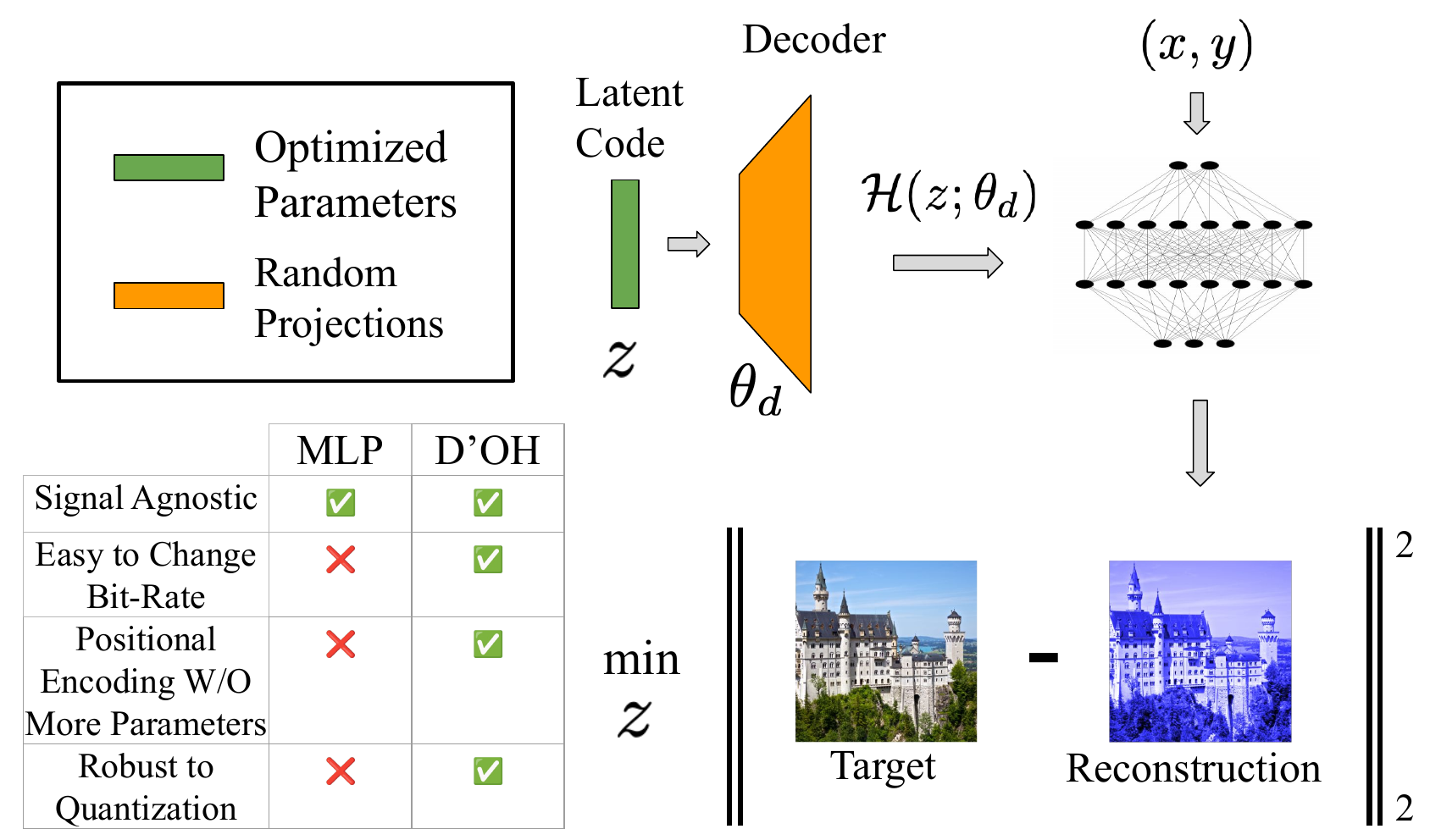}
    \caption{Illustration of the proposed Decoder-Only Hypernetwork, which optimizes a low-dimensional latent code $z$ to generate weights for the target implicit neural representation (INR). This signal-agnostic framework operates without offline training data, relying solely on the target architecture and the specific data instance. Random projections act as the network decoder $\theta_d$, facilitating a compact code representation.}
    \label{fig:decoder-banner}
\end{figure}

\begin{figure}[h!]
    \centering
    \includegraphics[width=1.0\linewidth]{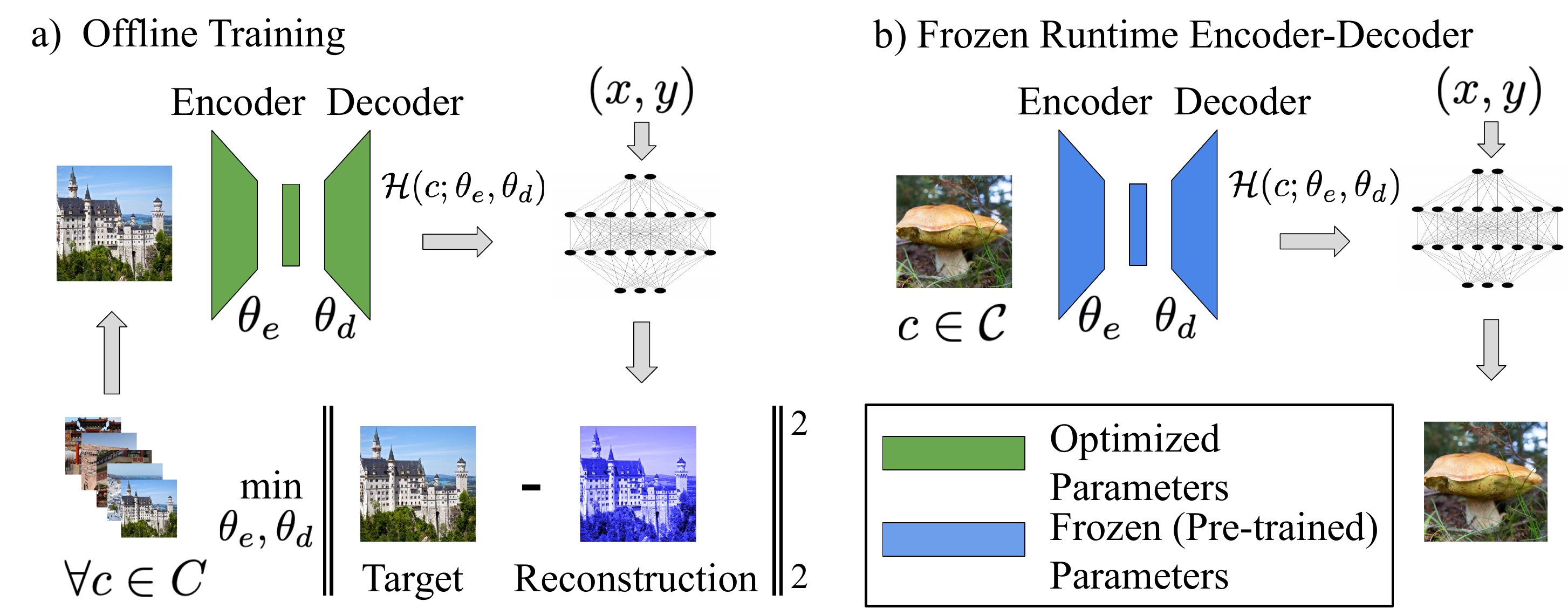}

    \caption{Conventional Encoder-Decoder Hypernetworks are optimized offline on a signal class. The hypernetwork [$\theta_e, \theta_d$] is frozen and runtime predicts INR weights for new data instances, limiting generality for out-of-distribution signals. In contrast, a Decoder-Only Hypernetwork (see: Figure \ref{fig:decoder-banner}) is runtime optimized using only the target instance.}

    \label{fig:encoder-decoder}
\end{figure}

 Neural networks are believed to contain significant parameter redundancy, motivating interest in the use of more compact architectures \cite{denil_predicting_2013,ha_hypernetworks_2016, hinton_distilling_2015, goodfellow_deep_2016}. Hypernetworks (meta-networks that generate the weights of a target network) are a popular approach to approximate the performance of more expressive architectures \cite{galanti_modularity_2020, ha_hypernetworks_2016}. Often they are pre-trained on a target signal class or application -- for example, super-resolution, in-painting, or style transfer \cite{alaluf_hyperstyle_2022,nguyen_fast_2022,klocek_hypernetwork_2019, chauhan_brief_2023}. These hypernetworks are offline \textbf{encoders} of the characteristics of the signal class, which are runtime \textbf{decoded} to the target network weights (see: Figure \ref{fig:encoder-decoder}). Often they require more complex architectures than the target network - using highly-parameterized transformers or graph neural networks to predict simple MLPs \cite{chauhan_brief_2023}. 
 By using external data these hypernetworks can generate new networks for a target class of signals. However, this comes at the cost of compactness, the need for offline optimization, and a reduced ability to generalise to out-of-distribution data instances -- primary motivations for the use of INRs \cite{li_neural_2021, pistilli_signal_2022, dupont_coin_2021}. 

Hypernetwork research in INRs has largely focused on pre-trained \textbf{encoder-decoder} hypernetworks, which use data instances as an input to produce a new INR. In contrast, we suggest that a \textbf{decoder-only} form of hypernetwork is possible in which the data instance appears only as the target output. These Decoder-Only Hypernetworks (D'OH) can be optimised directly to predict a given data instance by projecting to a target implicit neural network structure (see: Figure \ref{fig:decoder-banner}). Similar to standard INRs this requires no offline training on the target signal class and can instead be optimised at runtime. We provide a simple model involving a low-dimensional parameter vector projected by a fixed linear random mapping to construct a target SIREN network \cite{sitzmann_implicit_2020}. By controlling a latent dimension for the hypernetwork and transmitting the random seed used to reconstruct the mapping to transmit the signal, we can use fewer parameters than the target network architecture -- enabling its use in compression. In contrast to existing INR compression methods such as COIN \cite{dupont_coin_2021,strumpler_implicit_2022}, which requires search on low-rate architectures or aggressive quantization to reduce bit-rate, our method keeps the same target architecture and instead directly varies latent dimension to compress to different rates. Furthermore, as SIRENs are highly sensitive to initialization \cite{ramasinghe_beyond_2022}, and standard initializations must be adapted for hypernetworks \cite{chang_principled_2020}, we develop a novel initialization scheme for this approach.

\noindent We make the following contributions in this paper: 
\begin{enumerate}
    \item We introduce a signal-agnostic framework (\textbf{D'OH} - \textbf{D}ecoder-\textbf{O}nly random \textbf{H}ypernetworks) for the runtime optimization of implicit neural representations which requires optimization only on a target signal instance, with no need for offline training on an example set from the target signal class. 
    \item We provide a simple Decoder-Only Hypernetwork architecture based on a trainable latent code vector and fixed random projection decoder weights. As only the latent code, biases, and an integer random seed need to be communicated to reconstruct signals we show this is applicable to data compression. 
    \item We derive a novel hypernetwork initialization to match the layer-weight variances of SIREN networks, which modifies the distribution of the non-trainable random weights rather than the optimizable parameters. 
\end{enumerate}
Finally, we present an intriguing result in Section \ref{sec:quantizationsmoothing} that using a random linear hypernetwork induces quantization smoothing, leading D'OH to achieve similar performance for both Post-Training Quantization (PTQ) and Quantization Aware Training (QAT). This result, although not a central contribution for the paper, is of practical interest due to the negligible training overhead of PTQ \cite{gholami_survey_2021}.

\section{Background and Motivation}

\subsection{Implicit Neural Representation Compression} 
\label{implicit_neural_compression}

An implicit neural representation (INR) is a function $f_\psi(\textbf{x})\to \textbf{y}$ mapping coordinates $\textbf{x}$ to features $\textbf{y}$ where $\psi$ are parameter values, trained to closely approximate a target signal $g$ such that $\|f_\psi(\textbf{x}) - g(\textbf{x})\| \leq \epsilon$. INRs are \textit{signal-agnostic} and have been widely applied to represent signal types including images, sound, neural radiance fields, and sign distance fields \cite{sitzmann_implicit_2020,mildenhall_nerf_2020,xie_neural_2022, lu_compressive_2021,mescheder_occupancy_2019,strumpler_implicit_2022,dupont_coin_2022}. Recently, INRs have garnered significant interest for their potential in implicit compression \cite{dupont_coin_2021,xie_neural_2022}. A forward pass of a trained network produces a lossy reconstruction of an original signal instance \cite{dupont_coin_2021,dupont_coin_2022,strumpler_implicit_2022,damodaran_rqat-inr_2023, dupont_data_2022}. Further compression can be achieved by exploiting the high-degree of redundancy known to exist in neural network weights \cite{han_deep_2016,denil_predicting_2013,martinez_permute_2021,ha_hypernetworks_2016}. For example, by quantizing network weights \cite{dupont_coin_2021,dupont_coin_2022, strumpler_implicit_2022, lu_compressive_2021, shi_distilled_2022,chiarlo_implicit_2021,damodaran_rqat-inr_2023}, pruning \cite{chiarlo_implicit_2021, lee_meta-learning_2021}, inducing sparsity \cite{yuce_structured_2022,schwarz_meta-learning_2022}, using neural architecture search, variational methods \cite{schwarz_modality-agnostic_2023}, hash-tables \cite{takikawa_variable_2022, takikawa_compact_2023}, latent transformations \cite{zhang_enhanced_2023, ladune_cool-chic_2023,kim_c3_2023} and applying entropy compression \cite{strumpler_implicit_2022,dupont_coin_2022, gordon_quantizing_2023, ladune_cool-chic_2023,kim_c3_2023}.

\subsection{Neural Architecture Search} \label{sec:nas}

\begin{figure}[ht]
    \centering
    \begin{subfigure}[b]{0.48\textwidth}
        \centering
        \includegraphics[width=\textwidth]{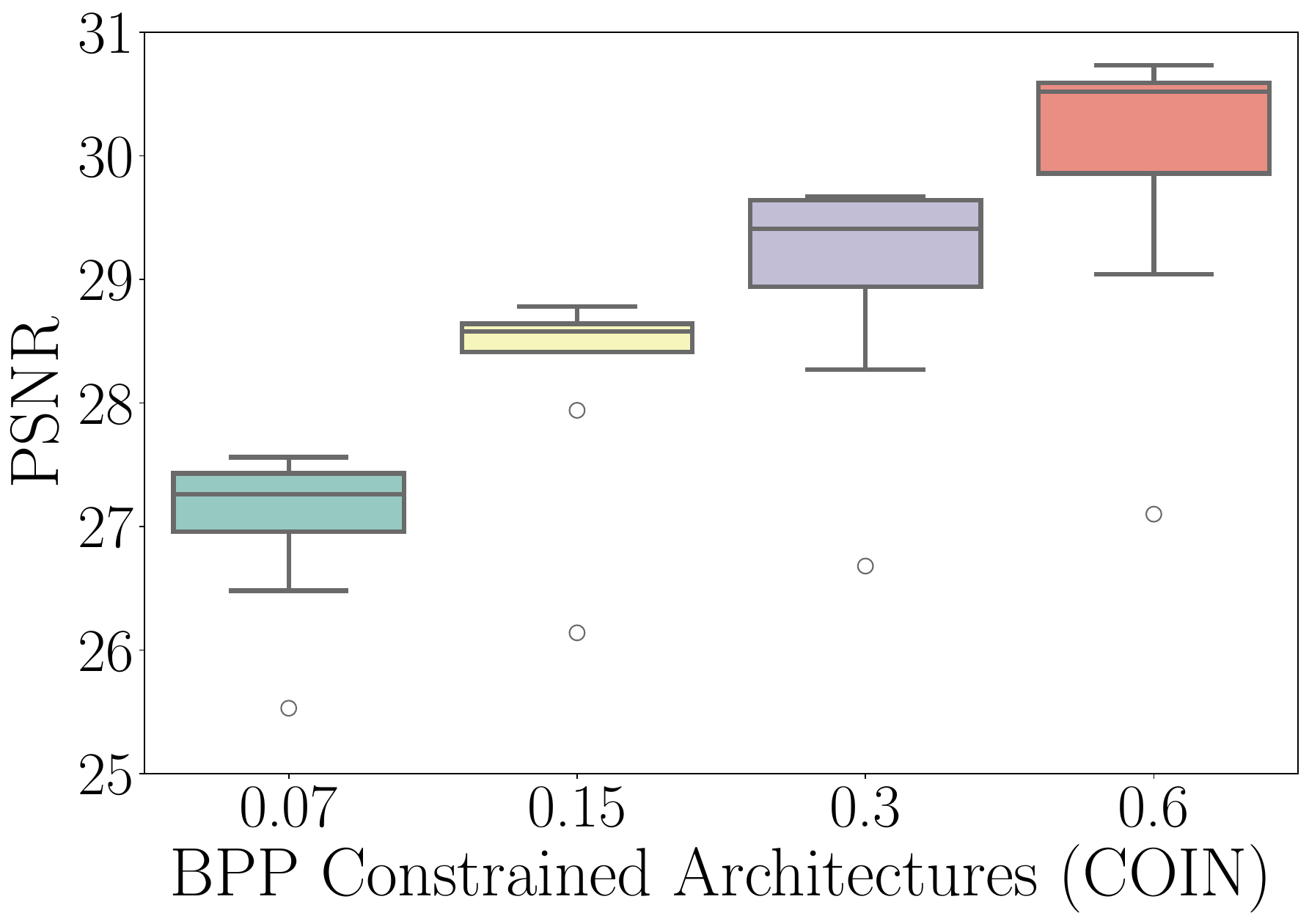}
        \label{fig:coin_psnr}
    \end{subfigure}
    \hfill %
    \begin{subfigure}[b]{0.48\textwidth}
        \centering
        \includegraphics[width=\textwidth]{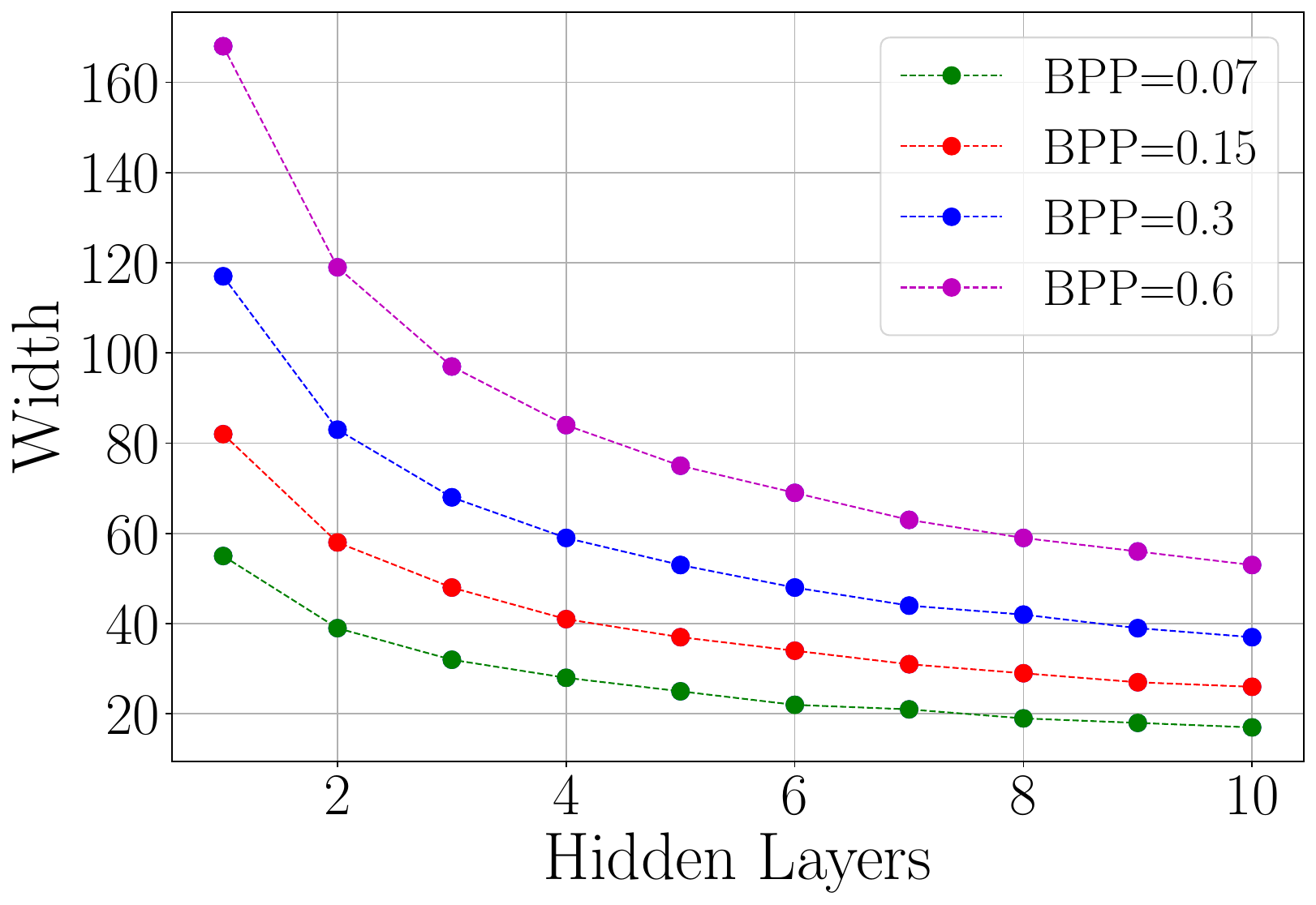}
        \label{fig:bpp_architectures}
    \end{subfigure}
    \caption{Left: COIN architectures show variability in outcome necessitating a costly Neural Architecture Search to achieve maximal performance as in \cite{dupont_coin_2021}. Right: Architectures that satisfy a bits-per-pixel constraint for COIN (Kodak index 2) using 16-bit quantization. There is a combinatorial increase in the search space for optimal architectures when considering different quantization levels (\eg for quantization aware training (QAT) as in \cite{strumpler_implicit_2022,gordon_quantizing_2023,damodaran_rqat-inr_2023} as QAT requires fixed quantization during training \cite{rastegari_xnor-net_2016}).}

    \label{fig:neuralarchitecturesearch}
\end{figure}

One of the difficulties in using INRs for compression is the need to decide between different architectures to achieve reductions in bit-rates \cite{vonderfecht2024predicting}. In COIN this is handled by performing Neural Architecture Search (NAS) \cite{elsken_neural_2019,white_neural_2023,dupont_coin_2021}. MLP architectures (width, hidden) that satisfy a bits-per-pixel (BPP) target (e.g. 0.3) on the Kodak dataset \cite{noauthor_true_1991} are searched over with the best performing models at each bit-rate selected for further experiments. Achieving similar bit-rates without a change in architecture would require aggressive quantization, which can severely degrade reconstruction accuracy. Figure \ref{fig:neuralarchitecturesearch} shows that NAS is useful for COIN, as architectures vary in performance at the same bit-rate. This search is costly, due to the time needed to train each model and the large number of satisfying architectures \cite{vonderfecht2024predicting}. In contrast, by working with a fixed target architecture generated by a Decoder-Only Hypernetwork we can control the desired bit-rate by directly varying the latent code dimension. This reduces the search across alternative networks to a $O(1)$ decision for a target bit-rate.

\subsection{Hypernetworks} 
\label{background_hypernetworks}

A hypernetwork is a meta-network that generates the parameters of another network, typically referred to as the target network \cite{ha_hypernetworks_2016,chauhan_brief_2023}. As described by Ha \etal \cite{ha_hypernetworks_2016}, this is achieved by approximating a target network's architecture with a low-dimensional latent code and applying a generating function. For example:
\begin{equation}\label{hyperneteqn}
    \mathcal{H}_\theta(z) = \psi
\end{equation}
where $z$ is a latent code, $\theta$ are the hypernetwork weights, and $\psi$ refers to the generated parameters of a target network function~$f_{\psi}$. For simplicity, we will use $W_l$ to refer to weights of the $l^{th}$ layer of a $L$-layer multi-layer perceptron, and $\psi$ to refer to parameters in a more generic mapping. Hypernetworks have been applied to various use-cases \cite{chauhan_brief_2023}, including compression \cite{gao_model_2021, karimi_mahabadi_parameter-efficient_2021,nguyen_fast_2022}, neural architecture search \cite{zhang_graph_2020}, image super-resolution \cite{ klocek_hypernetwork_2019, wang_computing_2022} and sound generation \cite{szatkowski_hypersound_2022}. There is no single dominant hypernetwork architecture and a wide number have been used in practice, including transformers; convolutional, recurrent, and residual networks; generative adversarial networks, graph networks, and kernel networks \cite{chen_transformers_2022, ha_hypernetworks_2016, gao_model_2021, sendera_hypershot_2022, sendera_general_2023, zhang_graph_2020, chauhan_brief_2023}. Hypernetworks have recently been applied to INRs, with the goal to generate task-conditioned networks suitable for use on new data instances \cite{chauhan_brief_2023}. Recent applications have included volume rendering \cite{wu_hyperinr_2023}, super-resolution of images \cite{klocek_hypernetwork_2019,wu_hyperinr_2023}, hyperspectral images \cite{zhang_implicit_2022}, sound \cite{szatkowski_hypersound_2022}, novel view synthesis \cite{wu_hyperinr_2023, chen_transformers_2022, sen_hyp-nerf_2023}, and partial-differential equations \cite{belbute-peres_hyperpinn_2021,majumdar_hyperlora_2023}.

\subsection{Encoder-Decoder Hypernetworks}

Hypernetworks as currently applied to INRs can be described as being \textit{encoder-decoders}, where an example set drawn from the target signal class is used to condition a hypernetwork before being evaluated on a target signal instance. Given a class $\mathcal{C}$ of a signal type, a training set $C \in \mathcal{C}$ with samples $c \in C$ is used to train a hypernetwork $\mathcal{H}(c;\theta_e,\theta_d)$ to generate the weights $\psi$ of the implicit neural function $f_\psi(x)$ where~$x$ is the input coordinate. The network is optimized offline to minimise the loss $\sum_{x} \|f_\psi(x) - c(x) \|^2_2$ across the training set, thereby learning characteristics of the signal class. At runtime the hypernetwork is used to generate the weights for an unseen data instance in the target class $c\in \mathcal{C}, c\notin C$ for use a downstream task. We have used $\theta_e, \theta_d$ to collectively define the parameters for encoding (learning from the data class) and decoding (generating the target network), but these can be part of a single hypernetwork architecture or more distinctly separated \cite{chauhan_brief_2023}. Examples of encoder-decoder hypernetworks for INRs include Klocek \etal \cite{klocek_hypernetwork_2019} who train on DIV2K to produce INRs for image super-resolution \cite{agustsson_ntire_2017}; Sitzmann \etal \cite{sitzmann_implicit_2020}, who train on CelebA for image in-painting \cite{garnelo_conditional_2018,liu_deep_2015}; Zhang \etal \cite{zhang_implicit_2022}, who focus on hyperspectral images; and Szatkowski \etal \cite{szatkowski_hypersound_2022}, who train on VCTK to generate audio INRs. While not directly using a hypernetwork, several related works have sought to condition a base network on target class data through meta-learning techniques such as MAML \cite{finn_model-agnostic_2017}, before using this as an initialization for fine-tuning, autoencoding, or learning a set of modulations for a novel data instance \cite{schwarz_modality-agnostic_2023,dupont_coin_2022,strumpler_implicit_2022,tancik_learned_2021,pham_autoencoding_2023}.

\section{Methodology}
\label{method}

\subsection{Decoder-Only Hypernetworks}
\label{decoder-only-hypernetworks}
\begin{figure}
    \centering
    \includegraphics[width=0.85\linewidth]{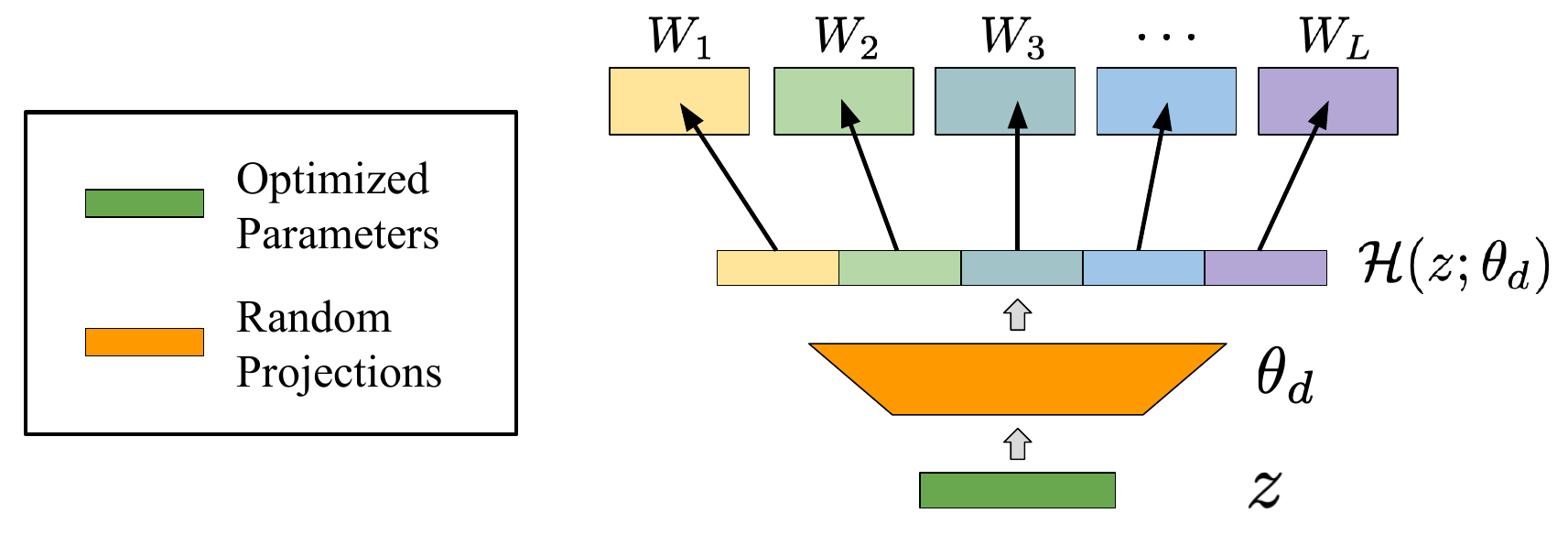}
    \caption{Weight generation in a Decoder-Only Hypernetwork. Latent code $z$ and decoding parameters $\theta_d$ are used to generate target network weights as the hypernetwork output $\mathcal{H}(z;\theta_d)$. We specifically investigate a random linear hypernetwork where $\mathcal{H}(z;\theta_d)=B_lz$ where $B_l$ is a fixed and untrained per-layer random weight matrix.}
    \label{fig:enter-label}
\end{figure}

While in Encoder-Decoder Hypernetworks the goal is to learn a domain-conditioned hypernetwork mapping to the target network, we instead propose to learn a latent code at runtime directly from the target data itself. As this process does not involve a pre-training stage of encoding domain information to the latent space, we describe this as a `Decoder-Only Hypernetwork' in which the latent code is optimized runtime by directly projecting to a target network. A Decoder-Only Hypernetwork is denoted as $\mathcal{H}(z;\theta_d)\to \psi$. For our purposes we restrict ourselves to the situation in which the decoder parameters $\theta_d$ are fixed and only the latent code $z$ is learned. As a specific example, we focus on a $L$-layer multi-layer perceptron target network and set the decoding weights to simply be \emph{linear maps} $\theta_d:=\{B_l\}_{l=1}^L$ defined by a family of \emph{fixed} random matrices $B_l$: one for each layer. The target network weights $W_l$ for the $l^{th}$ layer are defined to be the image of the latent vector $z\in \mathbb{R}^n$ under the linear map $B_l$. We maintain a separate trainable bias term $h_l$ for each layer in the target network. Denoting $\bar{W}_l$ as the generated vectorized form of $W_l$ Equation \eqref{hyperneteqn} reduces to:
\begin{equation}\label{eqn:W=BV}
    \mathcal{H}_\theta(z)=\{\bar{W}_l\}_{l=1}^L = \{B_l z\}_{l=1}^L.
\end{equation}

This simple random projection architecture has a number of advantages. Firstly, it exploits depth-wise redundancy in the target network by inducing parameter sharing \cite{nowlan_simplifying_1992, ullrich_soft_2016, savarese_learning_2019, chauhan_brief_2023}, as the optimized parameters are tied by arbitrary sampled random matrices. Secondly, a random matrix decoder may be transmitted with a single random seed enabling a highly compact transmission protocol. A similar integer seed protocol was recently proposed by \cite{kopiczko_vera_2024} in the context of Low-Rank Adaptation (LoRA). We note that more general decoder-only architectures could extend beyond the linear hypernetwork case we have proposed to include more expressive latent variables (\eg per-layer latent variables $z_l$), different fixed mappings, or non-linear decoders. The auto-decoder approach of Park \etal \cite{park_deepsdf_2019} can be considered an example using a fully-parameterised decoder. The effective use of random matrices in the low-rank matrix decompositions described by Denil \etal \cite{denil_predicting_2013}, classical results applying random projections for dimension reduction \cite{bingham_random_2001,dasgupta_elementary_2003,halko_finding_2011,johnson_extensions_1984,xie_survey_2018}, and recent works exploring compressibility of neural architectures using random projections motivate our approach\cite{kopiczko_vera_2024,arora_stronger_2018,aghajanyan_intrinsic_2021,ramanujan_whats_2020,mcdonnell_ranpac_2023}.

\subsection{Hypernetwork Training and Initialization} 
\label{background_hypernetwork_training_and_init}

Initializing hypernetworks is non-trivial as standard schemes (such as He \etal  \cite{glorot_understanding_2010} and Glorot \etal \cite{he_deep_2016}) do not directly translate when parameters are contained in a secondary network \cite{chang_principled_2020}. As a result, without correction the target network may experience exploding or vanishing gradients \cite{chang_principled_2020}, and important convergence properties are not guaranteed even under infinite-width hypernetworks \cite{littwin_infinite-width_2021}. Additionally, activations that are known to be sensitive to initialization, such as SIREN, require careful consideration even for standard MLPs \cite{ramasinghe_beyond_2022,sitzmann_implicit_2020}. The common principle for initialization schemes is to set weights so that layer-output variance is preserved following activation \cite{he_deep_2016, glorot_understanding_2010}. For hypernetworks we instead need to maintain the output variance for the target network. Chang \etal explore strategies for tanh and ReLU target networks by modifying parameter initialization in the hypernetwork. Linear hypernetworks as in Equation \eqref{eqn:W=BV} using per-layer random projections and a shared latent code suggests a different strategy -- changing the variance of the random projections. It can be shown\footnote{A full derivation and SIREN equivalent initialization is provided in Supplementary Materials \ref{Supplementary_derivation} 
}  that the variance of target network weights $\bar{W}_l=B_l z$ depends only on the variance of $B_l$, the variance of $z$, and the latent dimension $n$. We therefore set the random matrices to be uniformly initialized such that: 

\begin{equation}\label{eqn:varianceequation}
   \Var{ B_l} = \frac{\Var{\bar{W}_l}}{n\Var{z}}.
\end{equation} 

This ensures that the projected weight matrices will have identically distributed entries of the same variance as a given target network. In our experiments we initialize to preserve the variances of the original SIREN implementation \cite{sitzmann_implicit_2020}. As the SIREN input-layer is initialized separately to the rest of the network we would not be able to match all layers by initializing the latent code alone. We arbitrarily initialize the latent code uniformly between $[-1/n, 1/n]$. Biases are not tied to the latent code and are separately initialized to zero, excluding the output bias which is set to 0.5 in the middle of the output range.

\subsection{Training Configurations and Metrics}

We test the performance of D'OH on two implicit neural compression tasks: image regression and occupancy field representation. We choose a target network of 9 hidden layers and width 40, corresponding to the 0.6 BPP model for KODAK in \cite{dupont_coin_2021}. We test latent code dimensions calculated to represent  approximately 100\%, 60\%, and 30\% of the parameters of the target (9, 40) MLP model as calculated without positional encoding. Note that the number of parameters increases for MLP models with positional encoding, but not with D'OH (see: Section \ref{sec:positional_encoding} for discussion). For MLP baselines, we train models using the configurations in COIN at 0.07, 0.15, 0.3 BPP, corresponding to (width, hidden) pairs of (20,4), (30,4), (28, 9), and (40,9) \cite{dupont_coin_2021}. See Supplementary Table \ref{tab:training_configs} for the full set of configurations used for training. We experiment both with and without positional encoding (10 Fourier frequencies \cite{mildenhall_nerf_2020}). Following training we apply Post-Training Quantization to model weights at the best performing epoch and calculate the perceptual metrics at each quantization level. Compression metrics are reported using the estimated memory footprint (parameters $\times$ bits-per-parameter) and the bits-per-pixel (memory/pixels). We find this to be a close proxy to the performance of an entropy compressor (BZIP2 \cite{seward_bzip2_1996}), with some variation due to file overhead. See Supplementary Materials \ref{sec:quantization} for technical details of the applied Post-Training Quantization strategy and compression.

\begin{figure}[ht!]
    \centering
    \begin{subfigure}[b]{0.47\textwidth}
        \includegraphics[width=\linewidth]{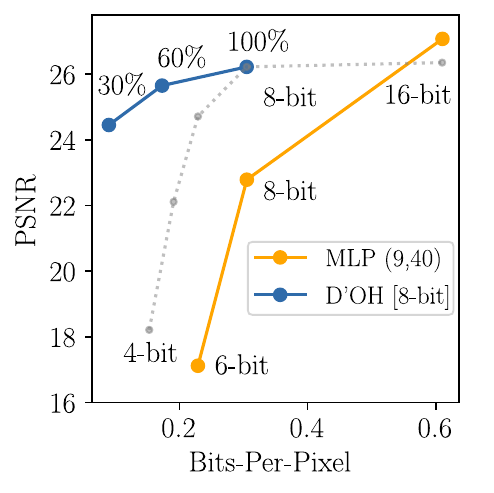}
        \caption{D'OH is more robust to quantization (grey) than MLPs (orange). As such we can pick an optimal quantization level (8-bit) and directly varies the latent code to control bit-rate (blue). }
        \label{fig:subfigure_quantization}
    \end{subfigure}
    \hfill 
    \begin{subfigure}[b]{0.48\textwidth}
        \centering
        \includegraphics[width=\linewidth]{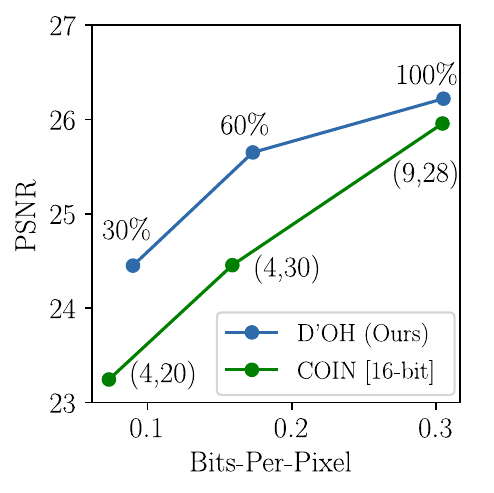}
        \caption{D'OH avoids Neural Architecture Search by maintaining the same target architecture, enabling a simple choice for a lower rate model. \\}
        \label{fig:subfigure_nas}

    \end{subfigure}

    \caption{Reducing bit-rate for a fixed MLP requires either a) Aggressive Quantization, or b) Neural Architecture Search on a space of low-rate structures. Kodak dataset.}
    \label{fig:quantization_subfigures}
\end{figure}

\section{Results}
\label{results}

\subsubsection{Image Compression - Kodak}
\begin{figure}
    \centering
    \includegraphics[width=\linewidth]{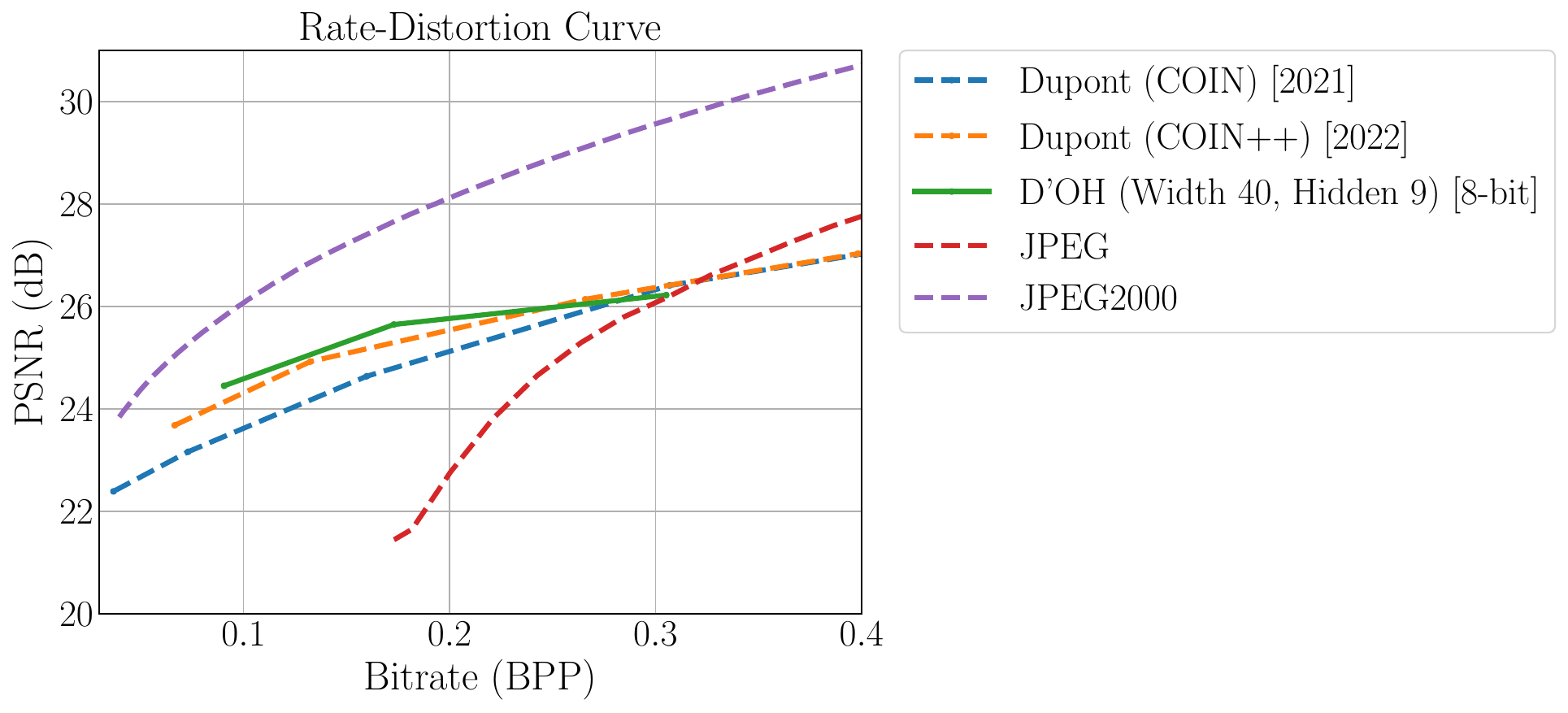}
    \caption{Rate-Distortion curve on the Kodak dataset. D'OH approximates a single target architecture (Width 40, Hidden 9) with 8-bit Post-Training Quantization. The rate distortion curve is generated by varying the latent code dimension. The D'OH model exceeds the performance of JPEG, COIN, and COIN++.}
    \label{fig:kodak_comparison}
\end{figure}

Image regression is a $\mathbb{R}^2:\mathbb{R}^3$ coordinate function used to implicitly represent 2D images by predicting RGB values at sampled coordinates. We conduct image compression experiments on the Kodak dataset \cite{noauthor_true_1991}. The Kodak dataset consists of natural images with dimensions of $768\times512$ pixels and is a common test for implicit neural image compression (e.g. \cite{dupont_coin_2021,dupont_coin_2022,strumpler_implicit_2022,damodaran_rqat-inr_2023}. We test across the 24 image Kodak dataset and report rate-distortion performance for the 100\%, 60\%, and 30\% latent code dimensions, using 8-bit PTQ. We select three literature Kodak benchmarks to highlight as examples of a signal specific codec (JPEG), a signal-agnostic and data-less codec (COIN) \cite{dupont_coin_2021}, and a meta-learned signal-agnostic codec (COIN++) \cite{dupont_coin_2022}; however, our most direct point of comparison is the original COIN as it relies on no external data or signal specific information. Figure \ref{fig:kodak_comparison} shows the performance of D'OH relative to comparison algorithms as reported by CompressAI \cite{begaint_compressai_2020}. D'OH shows improved rate-distortion performance relative to JPEG, COIN, and COIN++. In the Supplementary Materials \ref{sec:kodaksupplementary} we provide additional benchmarks showing performance relative to leading compression methods that use domain information, including those incorporating meta-learned Quantization Aware Training \cite{strumpler_implicit_2022}, auto-encoders \cite{balle_nonlinear_2020}, and advanced image codecs \cite{bellard_bpg_2015}. 

\noindent \textbf{Qualitative} Qualitative results are shown in Figure \ref{qualitative_results_comparison}. As COIN is fixed at 16-bits \cite{dupont_coin_2021}, we use an additional MLP benchmark to show direct comparison to D'OH at 8-bit quantization. We can note that D'OH greatly outperforms the 8-bit MLP models, potentially due to reduced quantization error when using a single latent code. This is consistent with Figure \ref{fig:subfigure_quantization}. When compared to COIN (16-bit), which requires different architectures to vary bit-rate, D'OH out-performs at all comparable bit-rates while using the same target architecture. Note that in this experiment we use positional encoding for the D'OH model (which uses no additional parameters), and no positional encoding for the MLP models. In Supplementary Figure \ref{fig:pe_combined} we demonstrate reduced rate-distortion performance for very low-rate MLP image models with positional encoding, due to increased input layer parameters, so this is a stronger benchmark comparison.

\subsubsection{Image Regression - DIV2K} 
\begin{figure}[h!]
    \centering
    \includegraphics[width=\textwidth]{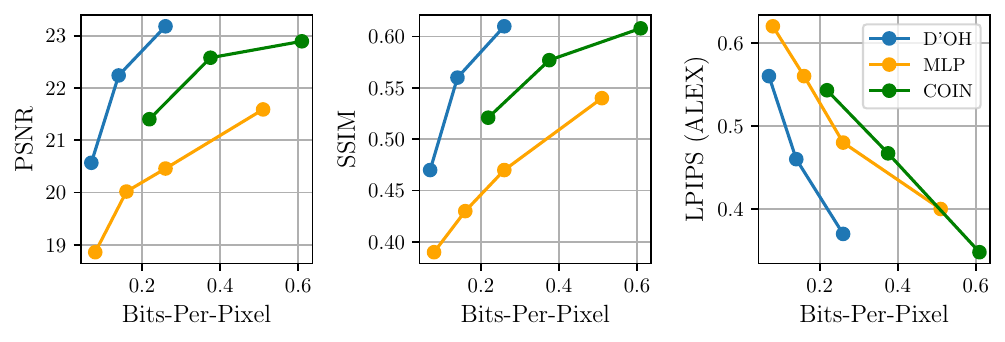}
    \caption{DIV2K Image Regression with 8-bit PTQ. Results averaged over 10 indices. D'OH shows improved performance relative to the MLP. This corresponds to a BD-Rate of -78.66\%, and BD-PSNR of 2.38db indicating substantial rate reduction, evaluated using the maximal performance at each rate and applying Akima interpolation\cite{bjontegaard_calculation_2001,barman_bjontegaard_2024}.} 
    \label{fig:div2k}
\end{figure}

DIV2K is a $512\times 512$ pixel natural image dataset frequently used for image regression \cite{agustsson_ntire_2017}. We repeat the experiment for 10 indices of the DIV2K dataset, using the same D'OH and MLP configurations from the Kodak experiment, training for 500 epochs per instance, and applying 8-bit PTQ. Figure \ref{fig:div2k} shows the results evaluated for PSNR$\uparrow$, Shared-Structural Similarity (SSIM$\uparrow$) \cite{wang_image_2004}, and Learned Perceptual Image Patch Similarity loss (LPIPS$\downarrow$) \cite{zhang_unreasonable_2018}. Results show that D'OH performs better than the MLPs across each metric. 

\subsubsection{Occupancy Field Experiments}

\begin{figure}[h!]

    \centering
    \includegraphics[width=\textwidth]{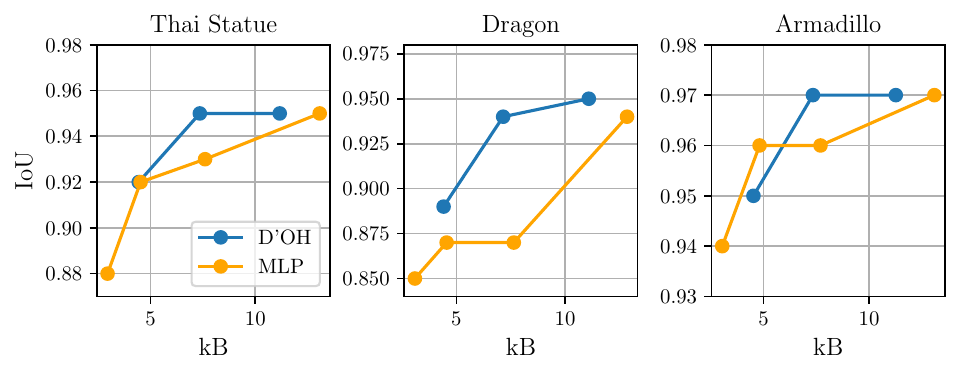}
    \caption{Binary Occupancy Field experiments (6-bit quantization). D'OH exhibits improved rate-distortion improvement over quantized MLPs. Rate-distortion generated for D'OH by varying the latent dimension and by varying architecture for MLPs.} 
    \label{fig:occupancy_6bit}

    \bigskip

    \centering
    \includegraphics[width=\textwidth]{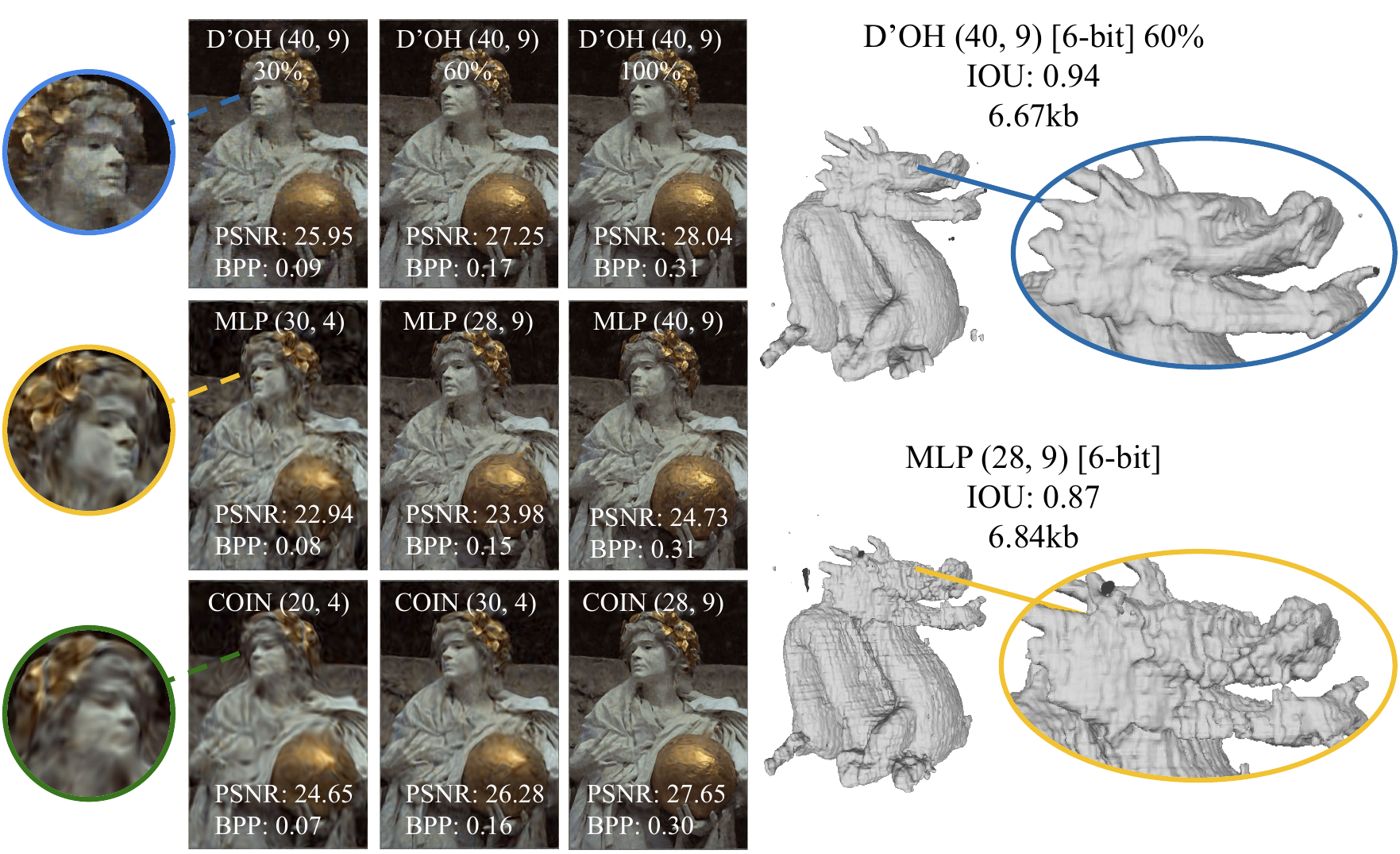}

    \caption{Qualitative image and occupancy field results. Left: Kodak. The top row corresponds to D'OH (8-bit quantization) with a target model (40 width, 9 hidden layers) with varied latent dimension. The second and third rows are MLPs (8-bit and 16-bit [COIN]) to match approximate bit-rates. Right: Binary Occupancy Field. D'OH is robust to quantization, enabling occupancy reconstruction at low quantization levels.}
    \label{qualitative_results_comparison}
\end{figure}

A Binary Occupancy Field is a $\mathbb{R}^3:\mathbb{R}^1$ coordinate function used to implicitly represent 3D shapes with the model output representing a prediction of voxel occupancy \cite{mescheder_occupancy_2019}. We test the ability of D'OH to represent occupancy fields by exploring the implementation provided by \cite{saragadam_wire_2023} using Thai Statue, Dragon, and Armadillo instances \cite{noauthor_stanford_nodate}. The occupancy training set is constructed by sampling voxels in a coordinate lattice, with an indicator function indicating whether the voxel is filled. As in the image experiments, we use a target network of (40, 9), and approximate this with smaller latent codes. We report performance using Intersection Over Union (IoU)$\uparrow$. Qualitative results and are generated by applying marching cubes across a thresholded set of sampled coordinates \cite{lorensen_marching_1987}, and are shown in Figure \ref{qualitative_results_comparison}. Positional encoding is required for MLP baselines to achieve suitable performance. In contrast to the image regression experiments, we find that we need to quantize to 6-bits or lower to see a clear rate-distortion improvement over MLPs with positional encoding (see: Figure \ref{fig:occupancy_6bit}). In the Supplementary Materials \ref{sec:supp_results}, we provide ablations showing the impact of quantization levels, target architectures, and positional encoding.

\section{Discussion}
\label{discussion}

  In the previous section we showed that a Decoder-Only Hypernetwork is able to compactly represent a signal using a small latent code and demonstrate its potential for signal agnostic compression by quantizing in a manner similar to COIN \cite{dupont_coin_2021}. Here we outline two interesting aspects of the model that occur from using a latent code and random projections to approximate a target network.

\begin{figure}[ht!]
    \centering
    \begin{subfigure}[b]{0.47\textwidth}
        \includegraphics[width=\linewidth]{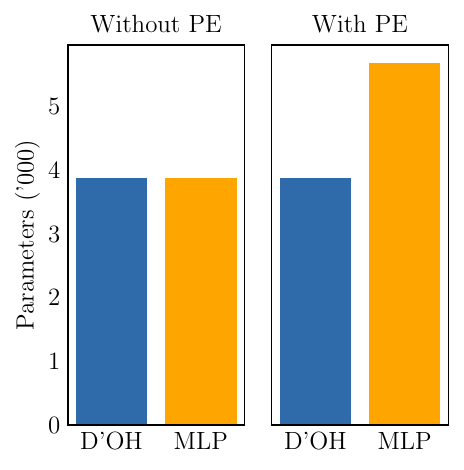}
        \caption{For very low small networks, the change in parameters induced by positional encoding can be significant: using a 10 frequency positional encoding for a 4 layer, 30 width network increases the number of MLP parameters by 46\%.}
        \label{fig:pe_param}
    \end{subfigure}
    \hfill 
    \begin{subfigure}[b]{0.48\textwidth}
        \centering
        \includegraphics[width=\linewidth]{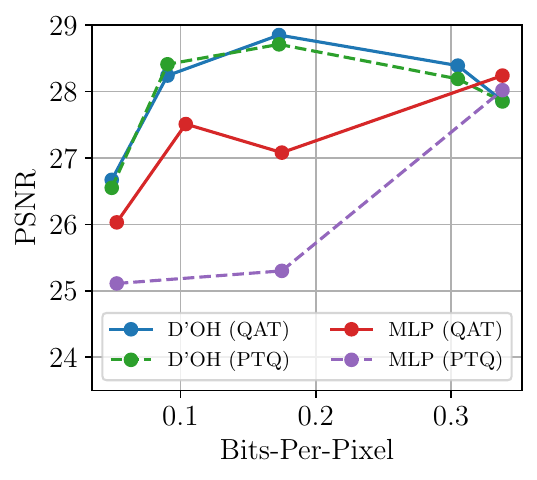}
        \caption{Without QAT, MLPs suffer from quantization error. In contrast, due to random projection smoothing, D'OH performs similarly for both QAT and PTQ for moderate quantization levels. Kodak index 12, 8-bit quantization.}
        \label{fig:QAT}
    \end{subfigure}
    \caption{The use of a random-linear hypernetwork for D'OH enables benefits beyond signal compression, including the use of positional encoding without an increase in model parameters and improved robustness to quantization without requiring QAT.}
    \label{fig:combined_figures}
\end{figure}

\subsubsection{Positional Encoding}\label{sec:positional_encoding}

The number of parameters used in the D'OH model is independent of the dimensions of the target network. An interesting consequence of this is that while including a \textit{positional encoding} component to a multi-layer perceptron increases the number of parameters (due to the increased input dimension), this is not true for D'OH. As a result the D'OH model is able to use positional encoding "for free". Figure \ref{fig:pe_param} shows for this can be significant for small networks. We find in Supplementary Figure \ref{fig:pe_combined} that the increase in parameters is sufficient to lead to reduced rate-distortion performance for small MLPs on image regression with positional encoding. In contrast, for the binary occupancy experiments we find that both the MLP and D'OH models noticeably improve in rate-distortion performance when applying positional encoding.

\subsubsection{Quantization Smoothing}
\label{sec:quantizationsmoothing}

\begin{figure}[htbp!]
    \centering
    \includegraphics[width=\linewidth]{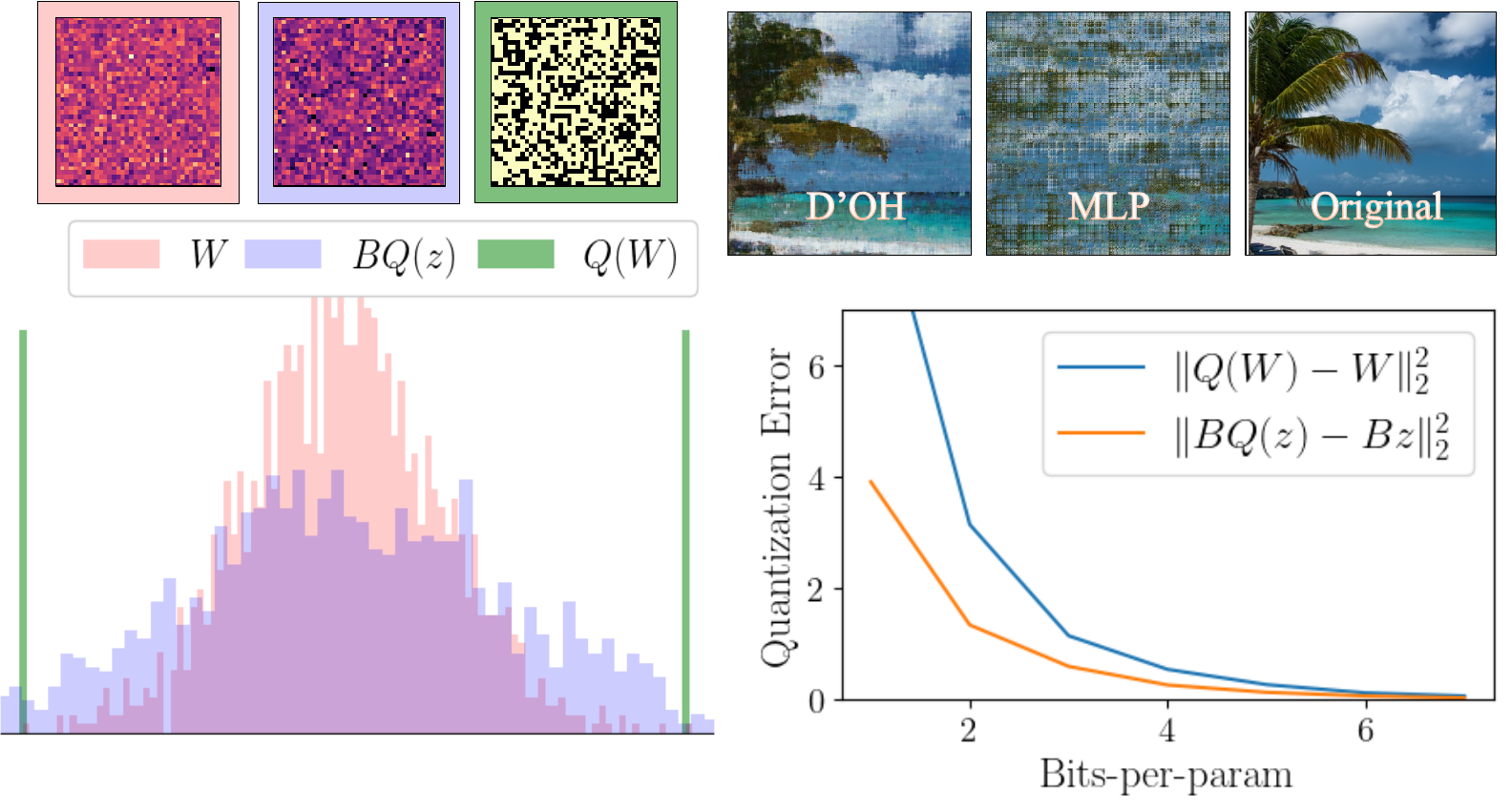}
    \caption{Random projection helps to smooth quantization artefacts. Left: direct 1-bit quantization \textcolor[rgb]{0.17,0.71,0.17}{$Q(W)$} of a matrix \textcolor[rgb]{0.99,0.34,0.34}{$W$} results in sharp weight discontinuities. Instead applying random projections to an quantized latent code \textcolor[rgb]{0.37,0.38,0.99}{$BQ(z)$} smooths values and better matches a unquantized distribution. Top-right: as a result D'OH can reconstruct images at 4-bit quantization, a rate which leads to catastrophic error for MLPs. Bottom-right: quantization error at different parameter quantization levels on a (40,40) matrix. }
    \label{fig:quantization}

\end{figure}

Despite using Post-Training Quantization for our experiments we observe that D'OH has lower quantization error relative to MLPs (see: Figure \ref{fig:subfigure_quantization}). This is likely due to the use of random projections in the hypernetwork. Directly applying a quantization map $Q$ to network weights $W$ leads to a large quantization error $\|Q(W)-W\|_2^2$ \cite{gersho_vector_1992}. In contrast for D'OH we quantize the latent code $z$ and \textit{only then} apply a per-layer full-precision random mapping $B$ to recover the target network weights. While the random projection adds no additional information it helps smooth sharp discontinuities in the weight space and allows the output distribution of $BQ(z)$ to better match the distribution of $Bz$. Figure \ref{fig:quantization} shows this visually - even with 1-bit quantization $BQ(z)$ can generate a full-distribution of weight values. Intriguingly, we find that D'OH performs similarly for both Post-Training Quantization and Quantization Aware Training (see Figure \ref{fig:QAT}), a result with practical application due to increased overhead and need to select fixed bit-rates when training with QAT \cite{gholami_survey_2021}.

\section{Limitations and Conclusion}
\label{conclusion}

 \subsubsection{Limitations} Although D'OH achieves significant quality improvements at low-rate image compression tasks (see: Figure \ref{qualitative_results_comparison}), we find that performance for binary occupancy fields is more limited. While D'OH is more robust at lower quantization levels ($\leq6$-bit) than MLPs with positional encoding, fully searching across architecture and quantization levels for MLPs can achieve similar rate-distortion performance to D'OH (see Supplementary Materials \ref{sec:supp_results}). Secondly, there remains the issue of sizing target architectures (we take the 0.6 COIN architecture as a baseline comparison). D'OH doesn’t fully remove the need to choose a target architecture, but it does provide a natural way to vary bit-rate for a fixed target architecture. In contrast, similarly varying bit-rate for quantized MLPs requires either architecture search for a completely new network or higher quantization levels (see: Figure \ref{fig:neuralarchitecturesearch}). In Supplementary Figure \ref{fig:ablationtargetarchitectures} we provide an ablation of target architectures showing that varying the latent code follows a smooth Pareto frontier, with larger targets dominating smaller ones for given bit-rates, partially alleviating this concern. Finally, D'OH shares a limitation common to INRs in training time. For example, fitting a Kodak image with 2000 epochs takes approximately half-an-hour per instance. Note that \cite{dupont_coin_2021} use 50,000 iterations per instance, and \cite{strumpler_implicit_2022} up to 25,000 for their non-meta-learned models. A full dataset evaluation on Kodak (24 images) therefore takes $\sim 12$h per rate configuration. 
 
 \subsubsection{Conclusion} In conclusion we have introduced a framework for direct optimization of a Decoder-Only Hypernetworks for INRs. Contrasting prior work applying hypernetworks to INRs, our method is data agnostic and does not require offline pretraining on a target signal class and is instead trained at runtime using only the target data instance. We have demonstrated the potential for a latent code and fixed random projection matrices to act as a Decoder-Only Hypernetwork, and shown that this improves upon contemporary methods for image compression such as COIN \cite{dupont_coin_2021}. Surprisingly, this random linear hypernetwork is observed to perform similarly when trained under QAT and PTQ under moderate quantization levels which may have practical applications beyond INRs.

 \subsubsection{\ackname} The authors wish to acknowledge Shin-Fang Ch'ng and Xueqian Li for their invaluable comments and discussion on the paper.

\bibliographystyle{splncs04}

\bibliography{ACCV/accv}

\clearpage 

\title{Supplementary Materials\\D'OH: Decoder-Only random Hypernetworks for Implicit Neural Representations}
\author{}
\institute{}
\authorrunning{C. Gordon, L. E. MacDonald, H. Saratchandran, S. Lucey}
\titlerunning{Supplementary Materials - Decoder-Only Hypernetworks (D'OH)}
\setcounter{page}{1}

\setcounter{section}{0}
\setcounter{figure}{0}
\setcounter{equation}{0}
\setcounter{footnote}{0}

\maketitle

\section{Initialization}
\label{Supplementary_initialization}

In Section \ref{background_hypernetwork_training_and_init} we noted that we need to apply a modified initialization scheme under the random matrix hypernetwork structure we examine. Under most intiialization schemes (e.g. He \cite{he_deep_2016}, Xavier \cite{glorot_understanding_2010},  and SIREN \cite{sitzmann_implicit_2020}), initialization is conducted separately for each layer - a property we want to preserve in the target network. However as we will use the \textit{same} latent parameter vector to generate each layer we will instead need to account for this by changing the per-layer random matrices to match the desired initialization of the target network.

\subsection{Derivation}
\label{Supplementary_derivation}

Assume the entries of $z$ are drawn independently and identically distributed from a distribution of variance $\Var{z}$, and that the weights of the $l^{th}$ layer of the target network are to have variance $\Var{W_l}$.  We seek a formula for the variance $\Var{B_l}$ of the distribution from which to independently and identically draw the entries of the random matrix $B_l$ such that the entries of $B_lz$ have variance $\Var{W_l}$.
We assume that \emph{all} entries for both $z$ and $B_l$ are drawn independently of one another, and with zero mean.
From Equation \ref{eqn:W=BV}, we have: 
\begin{equation}
    \bar{W}_l = B_l z.
\end{equation}
Recall that $n$ denotes the dimension of $z$, and use superscripts to denote vector and matrix indices.  Then the above equation can be written entry-wise as: 
\begin{equation}
    \Var{\bar{W}_l^{i}} = \Var{\sum\limits_{j=1}^n B_l^{ij}z^j}
\end{equation} 
Since the entries of $B_l$ and $z$ are all independent, we therefore have:
\begin{equation}
    \Var{\bar{W}_l^i} = \sum\limits_{j=1}^n \Var{ B_l^{ij}z^j}.
\end{equation} 
Again using independence of the entries of $B_l$ and $z$, we have: 
\begin{equation}
\begin{split}
    \Var{\bar{W}_l^i} &= \sum\limits_{j=1}^n \Var{ B_l^{ij}}\Var{z^j} \\& + \Var{ B_l^{ij}} \mathbb{E} (z^j)^2 + \mathbb{E}(B_l^{ij})^2 \Var{z^j},
\end{split}
\end{equation} 
which simplifies to:
\begin{equation}
    \Var{\bar{W}_l^i} = \sum\limits_{j=1}^n \Var{ B_l^{ij}}\Var{z^j}
\end{equation}
by our zero-mean assumption on the entries of $B_l$ and $z$.  Invoking our identically distributed assumption finally yields:
\begin{equation}
    \Var{\bar{W}_l} = n \Var{ B_l}\Var{z},
\end{equation}
so that:
\begin{equation}\label{varianceequation}
   \Var{ B_l} = \frac{\Var{\bar{W}_l}}{n\Var{z}}.
\end{equation}
We will use this formula to find bounds on a uniform distribution for $B_l$ in order to achieve the variance $\Var{W_l}$ of the weights considered in \cite{sitzmann_implicit_2020}. To initialize $B_l$ using a uniform distribution centred at 0, we must determine its bounds $\pm a$. Taking the variance of a uniform distribution, we have $\Var{B_l}=\frac{1}{12}(2a)^2 =\frac{a^2}{3}$. Substituting into Equation \eqref{varianceequation}, we have:
\begin{equation}
    \frac{a^2}{3}=\frac{\Var{\bar{W_l}}}{n \Var{z}},
\end{equation}
so that
\begin{equation}\label{eqn:uniform_var}a = \pm \sqrt{\frac{3\Var{\bar{W_l}}}{n \Var{z}} }.
\end{equation}

\subsection{SIREN Equivalent Initialization}

We can apply Equation \eqref{eqn:uniform_var} to derive an example SIREN initialization \cite{sitzmann_implicit_2020}. \\

\noindent \textbf{Input Layer}\footnote{We follow the SIREN initialization scheme provided in the Sitzmann et al. (2020) codebase, as this has been noted by the authors to have improved performance \cite{sitzmann_implicit_2020}}: Assume $z$ is initialized using $U \sim (\pm \frac{1}{n})$ and $\bar{W}_0$ by $U \sim (\pm \frac{1}{fan_{in}})$ where $fan_{in}$ represents the input dimension of the target network:
\begin{equation}
    \Var{\bar{W_0}}=\frac{1}{12}(\frac{2}{fan_{in}})^2=\frac{1}{3 fan_{in}^2}
\end{equation} 
\begin{equation}
    \Var{z}=\frac{(2/n)^2}{12}=\frac{1}{3n^2}
\end{equation}
\begin{equation}
    \Var{B_0}=\frac{\Var{\bar{W_i}}}{n \Var{z}}=\frac{1/(3fan_{in}^2)}{n /(3n^2)} = \frac{n}{ fan_{in}^2 }
\end{equation}
\begin{equation}\label{input}
    a_0 = \pm \sqrt{\frac{3n}{fan_{in}^2 }}
\end{equation}
\textbf{Other Layers}: $\bar{W_i}$ initialized using $U \sim (\pm \frac{1}{\omega\sqrt{h}})$, where $h$ refers to the number of hidden units, and $\omega$ the SIREN frequency. 
\begin{equation}
    \Var{\bar{W_i}}=\frac{1}{12}(\frac{2}{\omega\sqrt{h}})^2=\frac{1}{3\omega^2 h}
\end{equation}
\begin{equation}
\Var{B_i}=\frac{\Var{\bar{W_i}}}{n \Var{z}}=\frac{1/(3\omega^2 h)}{n /(3n^2)} = \frac{n}{\omega^2 h}
\end{equation}
\begin{equation}\label{eqn:hiddenlayer}
    a_i = \pm \sqrt{\frac{3n}{\omega^2 h}}
\end{equation}

\subsubsection{Numerical Comparison} 
\begin{figure}
    \centering
    \includegraphics[width=0.8\linewidth]{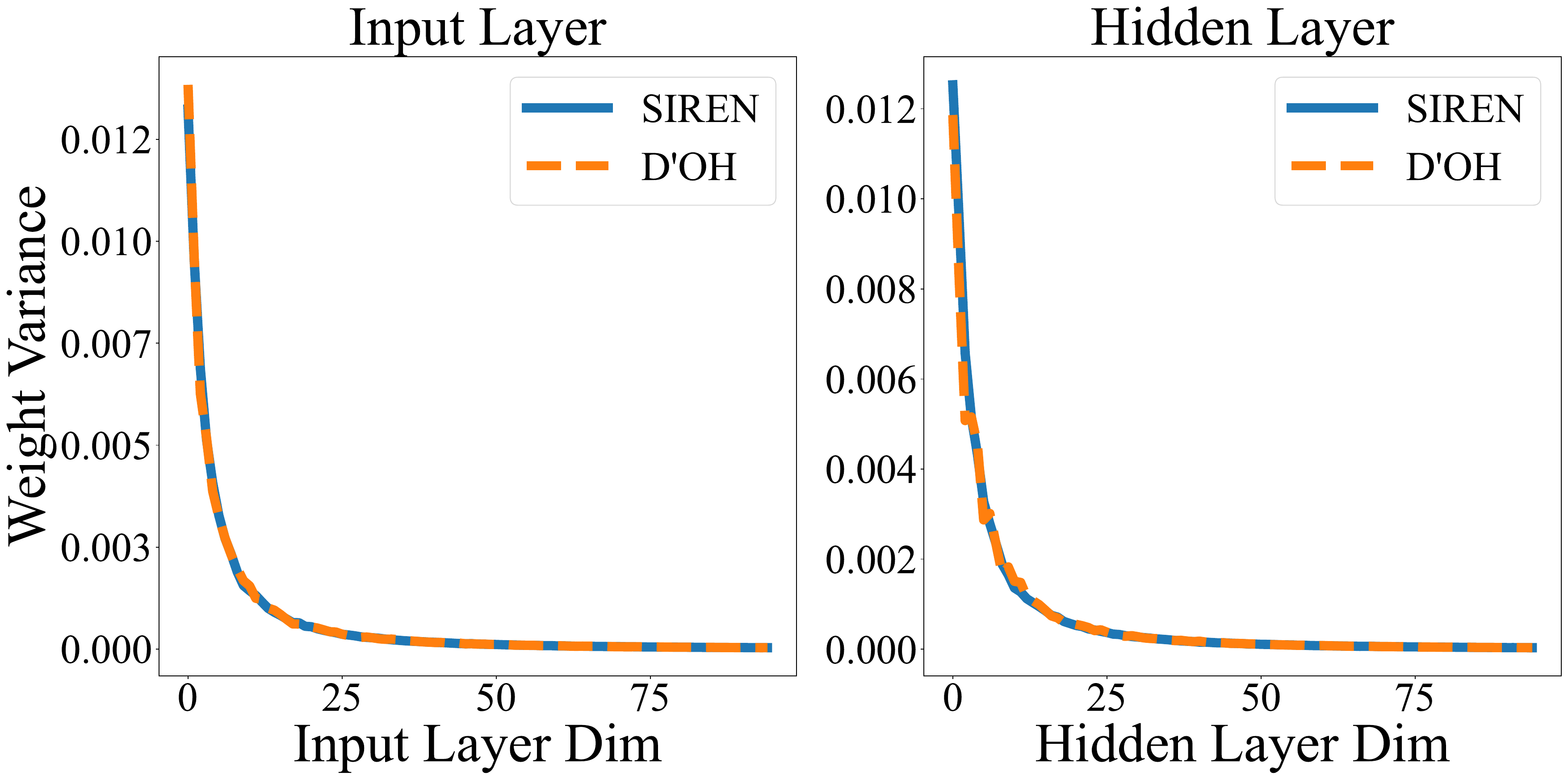}
    \caption{Numerical comparison of layer variances between SIREN and the weights generated by D'OH (latent dim: 2000 and $\omega=30$). Our initialization closely matches the initialization of SIREN \cite{sitzmann_implicit_2020}. }
    \label{fig:initialization}
\end{figure}

We initialize target networks with using Equations \eqref{input} and \eqref{eqn:hiddenlayer} for a range of input and hidden layer dimensions. The D'OH initialization correctly matches the target SIREN weight variances (Figure \ref{fig:initialization}).

\section{Quantization, Compression, and Transmission}

\subsubsection{Quantization}\label{sec:quantization} We outline here the design decisions for our quantization approach. We employ post-training quantization in our pipeline. While quantization-aware training (QAT) \cite{rastegari_xnor-net_2016} has been demonstrated to reduce quantization error in the context of implicit neural representations \cite{strumpler_implicit_2022, dupont_coin_2022, damodaran_rqat-inr_2023, gordon_quantizing_2023}, we note this has two key disadvantages: each quantization level needs to be trained separately, while post-training quantization can evaluate multiple quantization levels at the same time; and when quantization level is considered as part of the neural architecture search (see: Figure \ref{fig:neuralarchitecturesearch}) this expands the search space of satisfying models considerably. In addition, we employ a layer-wise range-based integer quantization scheme between the min and maximum values for each weight and distribution \cite{gholami_survey_2021}. We select an integer scheme to reduce the quantization symbol set \cite{gholami_survey_2021,gersho_vector_1992,jacob_quantization_2018}. We decided on a uniform quantization scheme rather than a non-linear quantizer such as k-means \cite{han_deep_2016} due to the overhead of code-book storage, which for small networks can be substantial proportion of compressed memory \cite{gordon_quantizing_2023}. In contrast, we represent each tensor with just three per-tensor components (integer tensor, minimum value, maximum value). Similar range-based integer quantization schemes are commonly described \cite{gholami_survey_2021,jacob_quantization_2018,krishnamoorthi_quantizing_2018}, and the method we use is only a subtle variation avoiding the explicit use of a zero point.

\subsubsection{Compression and Transmission} 

In a typical compressed implicit neural network the entire trained and compressed network weights need to be transferred between parties. This is done by first quantizing the weights followed by a lossless entropy compressor, such as BZIP2 \cite{seward_bzip2_1996} or arithmetic coding \cite{strumpler_implicit_2022}. Our method generates a target network by a low-dimensional linear code and fixed per-layer random matrices. As random matrices can be reconstructed by the transfer of an integer seed, we only quantize and compress the linear code. The recently proposed VeRA incorporates a similar integer seed transmission protocol for random matrices to improve the parameter efficiency of Low-Rank Adaptive Models \cite{kopiczko_vera_2024}. 

\begin{figure}
    \centering
    \includegraphics[width=0.8\textwidth]{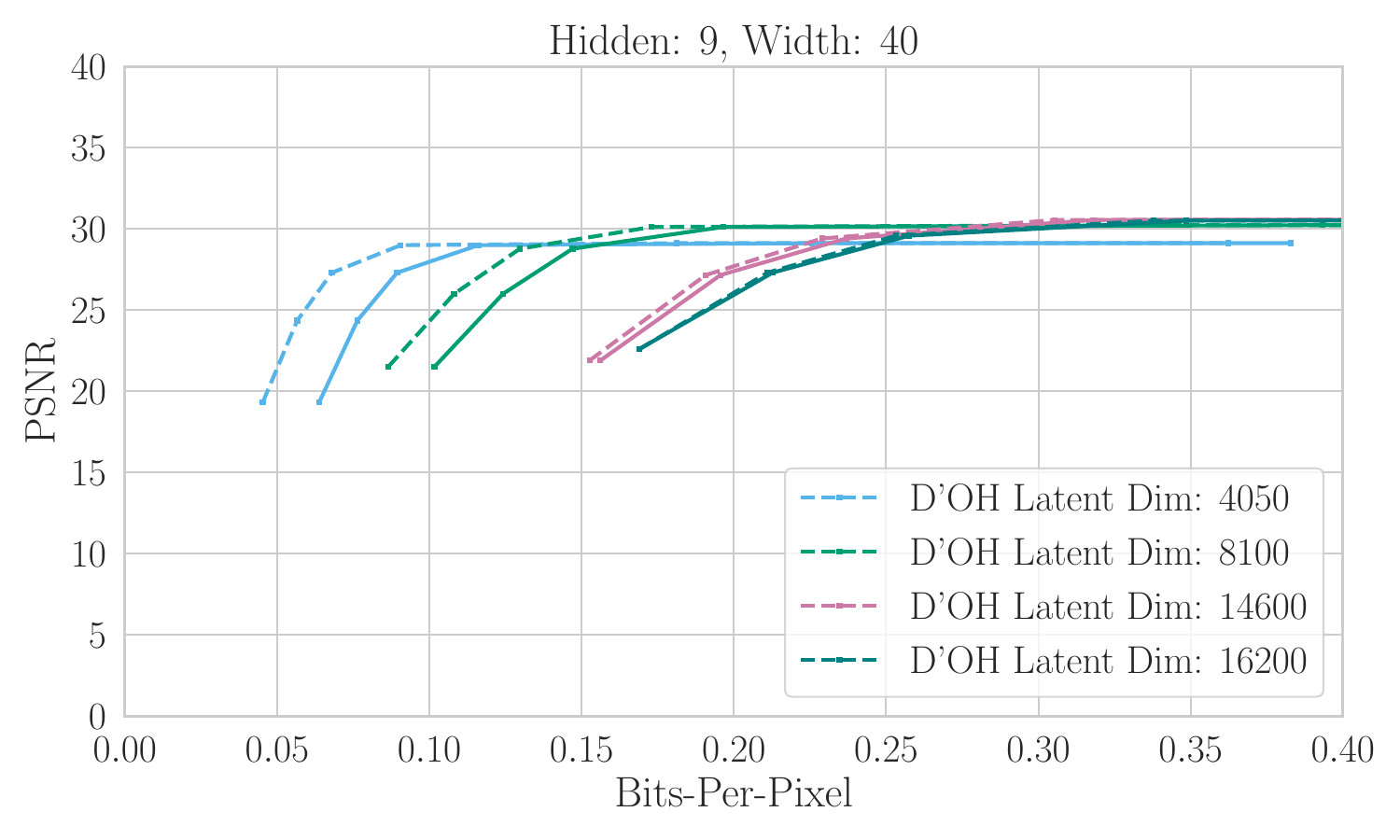}
    \caption{Comparison of bits-per-pixel (BPP) for estimated memory footprint (parameters $\times$ bits-per-weight) [dotted] and memory after applying BZIP2 [solid] to a Python pickle of the quantized model. Rate-distortions generated by varying quantization level. The estimated is a close proxy to an actual entropy coder, but shows some discrepancy at low-rate and low-quantization levels where file overhead represent a larger proportion of code size. To account for this we report the estimated memory footprint for both D'OH and MLPs, which can be seen as a overhead-free limit for performance.}
    \label{fig:bzip_comparison}
\end{figure}

\section{Positional Encoding}
\vspace{-15pt}
\begin{figure}[ht!]
    \centering
    \begin{subfigure}[b]{0.47\textwidth}
        \includegraphics[width=\linewidth]{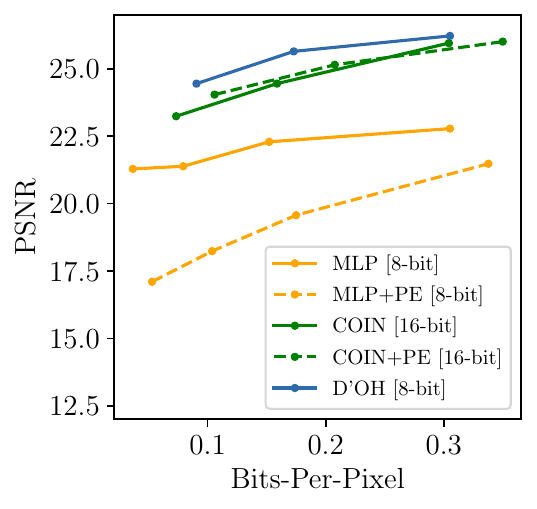}
        \caption{For image experiments we find that positional encoding reduces rate-distortion performance for MLPs at 8-bit, with little change observed at 16-bits. This is likely due to the increase in parameters and interaction with quantization effects. As a result we report image benchmarks without MLP positional encoding, as the stronger benchmark. Kodak.}
        \label{fig:pe_image}
    \end{subfigure}
    \hfill 
    \begin{subfigure}[b]{0.48\textwidth}
        \centering
        \includegraphics[width=\linewidth]{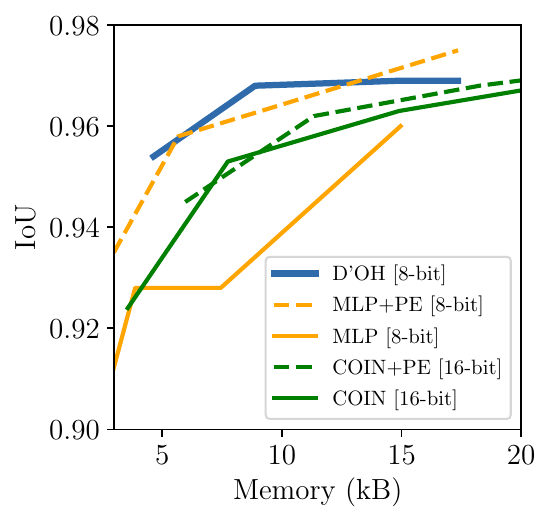}
        \caption{For Binary Occupancy experiments, we find that positional encoding is necessary for MLPs to obtain good reconstruction. This is possibly due to the presence of high-frequency spatial components in the 3D shape. Thai Statue. \\ \\ \\}
        \label{fig:pe_occupancy}
    \end{subfigure}
    \caption{Effects of positional encoding on Image and Binary Occupancy Experiments. D'OH does not increase parameters when using positional encoding (See: \ref{fig:pe_param}).}
    \label{fig:pe_combined}
\end{figure}

\vspace{-20pt}

\begin{table}[h!]
\centering
\caption{Training configurations for Image and Occupancy Field experiments.}
\begin{tabular}{|p{0.35\linewidth} | p{0.33\linewidth}|p{0.3\linewidth}|}
\hline
\textbf{Dataset} & \textbf{Images} & \textbf{Occupancy} \\ \hline
   Dimensions & Kodak $768 \times 512$ \newline  DIV2K $512\times 512$ & $512\times 512\times 512$ \\ \hline
Hardware & NVIDIA A100 & NVIDIA A100 \\ \hline
Optimizer & Adam $\beta=(0.99,0.999)$ & Adam $\beta=(0.99,0.999)$ \\ \hline
Scheduler (Exponential) & $\gamma=0.999$ & $\gamma=0.999$ \\ \hline
Epochs & 2000 & 250 \\ \hline
Batch Size & 1024 & 20000 \\ \hline
Loss & Mean Square Error & Mean Square Error \\ \hline
Perceptual Metrics & PSNR & IOU \\ \hline
Compression Metrics & Bits-Per-Pixel (BPP) & Memory (kB) \\ \hline
Target MLPs: width/hidden & 20/4, 30/4, 28/9, 40/9 & 20/4, 30/4, 28/9, 40/9 \\ \hline
Positional Encoding & 10 frequencies & 10 frequencies \\ \hline
Activation & Sine ($\omega=30$) & Sine ($\omega=30$) \\ \hline
Learning Rates (MLP/DOH) &  $2{e-4}$,$1e{-6}$ & $1{e-4}$, $1{e-6}$ \\ \hline
Quantization levels & [4, 5, 6, 8, 16] & [4, 5, 6, 8, 16] \\ \hline
\end{tabular}

\label{tab:training_configs}
\end{table}

\clearpage


\section{Further Benchmarks and Results}
\label{sec:supp_results}
\subsection{Kodak}
\label{sec:kodaksupplementary}
\begin{figure}
    \centering
    \includegraphics[width=0.9\textwidth]{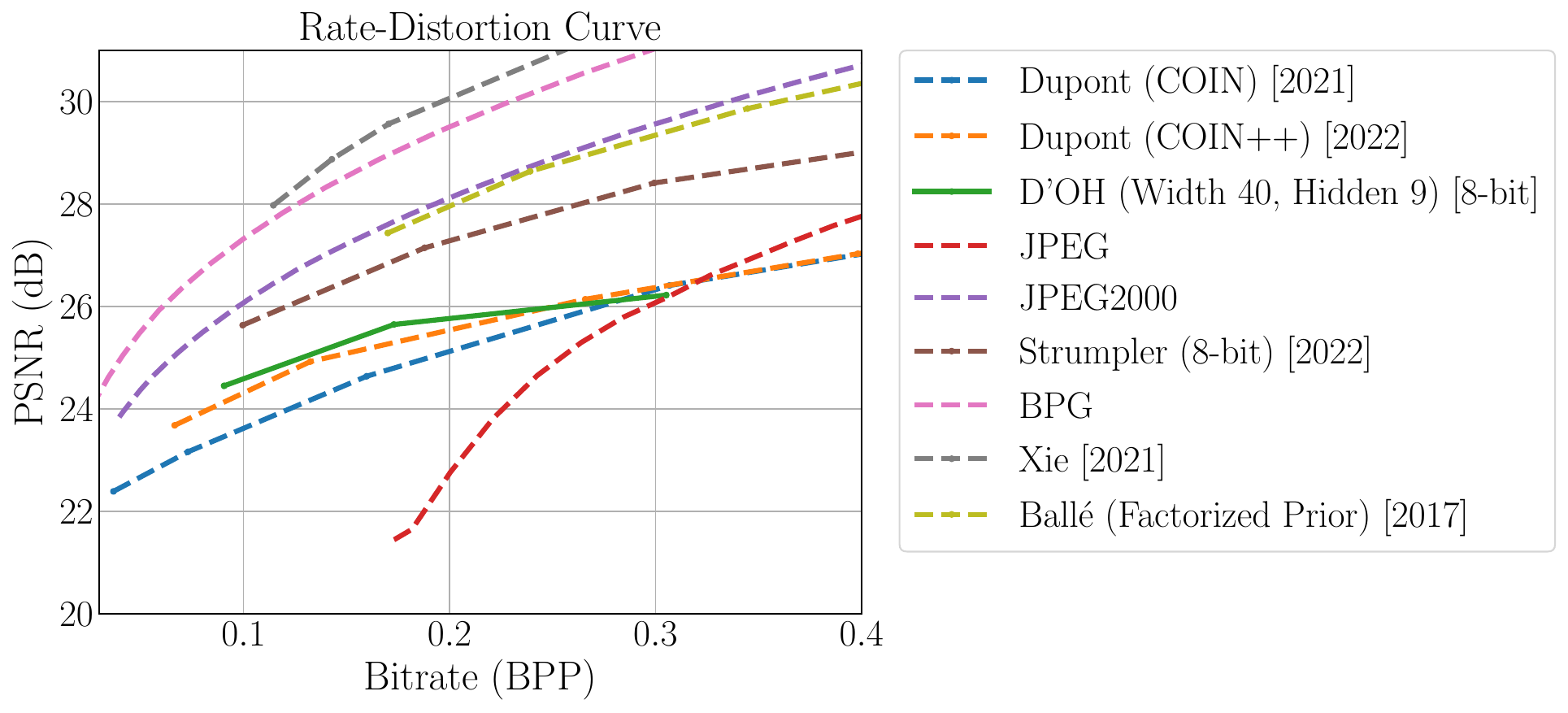}
        \caption{Rate-Distortion on Kodak showing additional benchmarks. Our method outperforms signal agnostic codecs trained without external datasets (COIN), our method lags both advanced signal specific codecs (JPEG2000 and BPG \cite{bellard_bpg_2015}), and those that employ auto-encoding \cite{balle_end--end_2017}, invertible encoding networks \cite{xie_enhanced_2021}, and meta-learned initializations \cite{strumpler_implicit_2022}. We suspect that the gap with \cite{strumpler_implicit_2022} is due to the use of quantization aware training (QAT). As mentioned in Section \ref{sec:nas}, we avoid QAT as a primary motivation for our method is to reduce the need for architecture search, including different quantization levels (the post-training quantization strategy we employ avoids this).}
    \label{fig:rate_distortion_8_bit_bench}
\end{figure}

\begin{figure}[h!]
    \centering
    \includegraphics[width=0.9\textwidth]{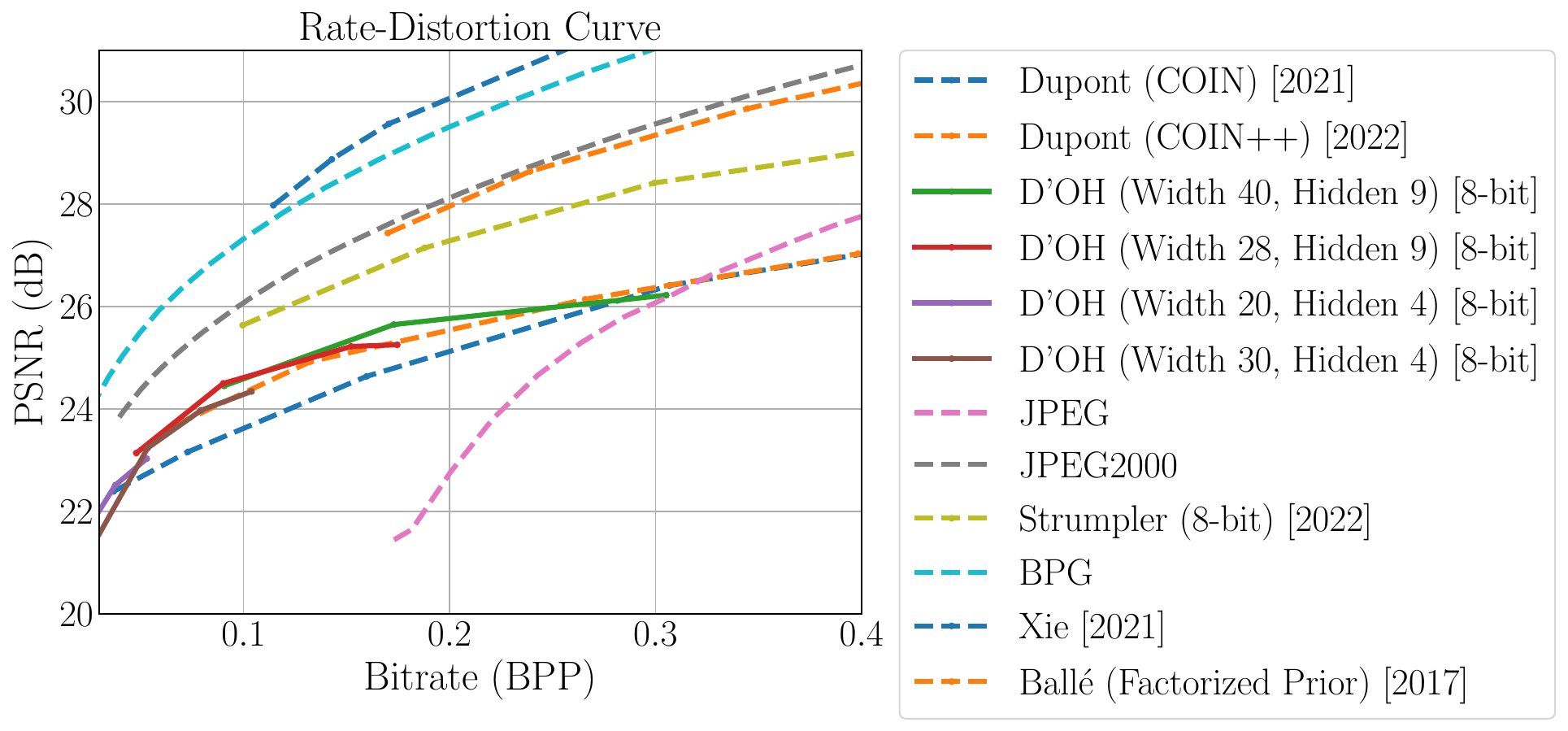}
    \caption{Ablation running D'OH with alternative COIN target networks. We note that D'OH is able to achieve a rate-distortion improvement on each of these architectures. The resulting model overlay shows an indicative Pareto frontier of the method. }
    \label{fig:ablationtargetarchitectures}
\end{figure}

\clearpage
\subsection{Occupancy Field}

\begin{figure}[h!]
    \centering
    \includegraphics[width=0.9\linewidth]{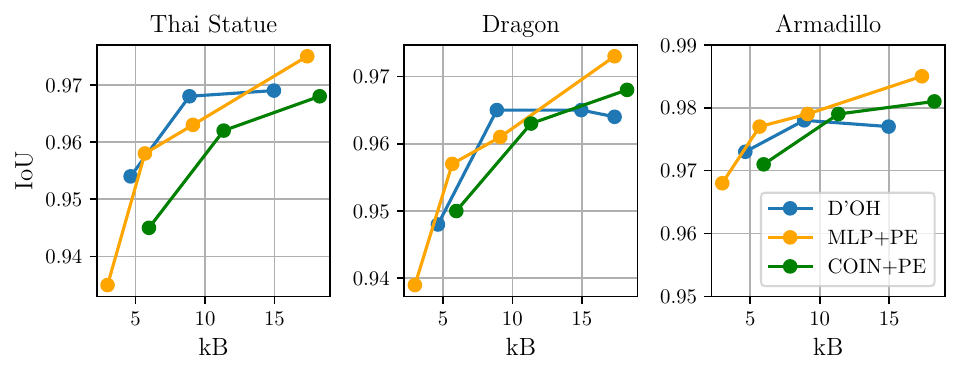}
    \caption{Rate-distortion curves for Binary Occupancy Fields on Thai Statue, Dragon, and Armadillo. D'OH and MLP are quantized at 8-bit and COIN at 16-bit. As positional encoding is required for Binary Occupancy performance (see: Supplementary Figure \ref{fig:pe_occupancy}, we apply it as the stronger benchmark. While a large rate-distortion advantage over MLPs is observed at 6-bit quantization (see: Supplementary Table \ref{tab:BO_all}), when evaluated across all quantization levels and architecture D'OH shows smaller improvement or close performance to MLP models with positional encoding.}
    \label{fig:occupancy_8bit}
\end{figure}

\clearpage

\begin{table}[h!]
\centering
\caption{Binary Occupancy Results for Thai Statue, Dragon, and Armadillo. D'OH performs substantially better than MLP models at low quantization levels (6-bit or lower), and MLPs without positional encoding. At higher quantization levels performance between MLPs and D'OH is comparable, with some rate distortion improvement observed for the 60\% D'OH. COIN represents a MLP with 16-bit quantization \cite{dupont_coin_2021}. }
\begin{tabular}{lccccc}
\toprule
&&\textbf{Memory}&& \textbf{IoU$\uparrow$} \\
\textbf{Model} & \textbf{Params} & \textbf{(kB)} & \textbf{Thai Statue} & \textbf{Dragon} & \textbf{Armadillo} \\ \midrule
\textit{6-bit} \\
MLP (4,20) & 1781 & 1.34 & 0.74 & 0.66 & 0.74 \\
MLP (4,30) & 3871 & 2.90 & 0.70 & 0.66 & 0.87 \\
MLP (9,28) & 7449 & 5.59 & 0.80 & 0.67 & 0.86 \\ 
MLP (9,40) & 14961 & 11.22 & 0.82 & 0.72 & 0.88 \\ \\ 
MLP+PE (4,20) & 2981 & 2.24 & 0.88 & 0.85 & 0.94 \\
MLP+PE (4,30) & 5671 & 4.25 & 0.92 & 0.87 & 0.96 \\
MLP+PE (9,28) & 9129 & 6.85 & 0.93 & 0.87 & 0.96 \\
MLP+PE (9,40) & 17361 & 13.02 & 0.95 & 0.94 & 0.97 \\ \\
DOH (30\%) & 4641 & 3.48 & 0.92 & 0.89 & 0.95 \\
DOH (60\%) & 8881 & 6.67 & 0.95 & 0.94 & 0.97 \\
DOH (100\%) & 14961 & 11.22 & 0.95 & 0.95 & 0.97 \\
\\
\textit{8-bit} \\
MLP (4,20) & 1781 & 1.78 & 0.89 & 0.85 & 0.94 \\
MLP (4,30) & 3871 & 3.87 & 0.93 & 0.87 & 0.96 \\
MLP (9,28) & 7449 & 7.45 & 0.93 & 0.88 & 0.96 \\
MLP (9,40) & 14961 & 14.96 & 0.96 & 0.93 & 0.97 \\ \\
MLP+PE (4,20) & 2981 & 2.98 & 0.94 & 0.94 & 0.97 \\
MLP+PE (4,30) & 5671 & 5.67 & 0.96 & 0.96 & 0.98 \\
MLP+PE (9,28) & 9129 & 9.13 & 0.96 & 0.96 & 0.98 \\
MLP+PE (9,40) & 17361 & 17.36 & 0.98 & 0.97 & 0.99 \\ \\
DOH (30\%) & 4641 & 4.64 & 0.95 & 0.95 & 0.97 \\
DOH (60\%) & 8881 & 8.88 & 0.97 & 0.97 & 0.98 \\
DOH (100\%) & 14961 & 14.96 & 0.97 & 0.97 & 0.98 \\
\\
\textit{16-bit} \\
COIN (4,20) & 1781 & 3.56 & 0.92 & 0.88 & 0.97 \\
COIN (4,30) & 3871 & 7.74 & 0.95 & 0.90 & 0.98 \\
COIN (9,28) & 7449 & 14.90 & 0.96 & 0.94 & 0.98 \\
COIN (9,40) & 14961 & 29.92 & 0.98 & 0.97 & 0.99 \\ \\
COIN+PE (4,20) & 2981 & 5.96 & 0.95 & 0.95 & 0.97 \\
COIN+PE (4,30) & 5671 & 11.34 & 0.96 & 0.96 & 0.98 \\
COIN+PE (9,28) & 9129 & 18.26 & 0.97 & 0.97 & 0.98 \\
COIN+PE (9,40) & 17361 & 34.72 & 0.98 & 0.98 & 0.99 \\
\bottomrule
\label{tab:BO_all}
\end{tabular}
\end{table}

\clearpage

\noindent\begin{minipage}{\textwidth}
    \subsection{Additional Qualitative Results - Kodak}

    \centering
    \includegraphics[width=0.9\textwidth]{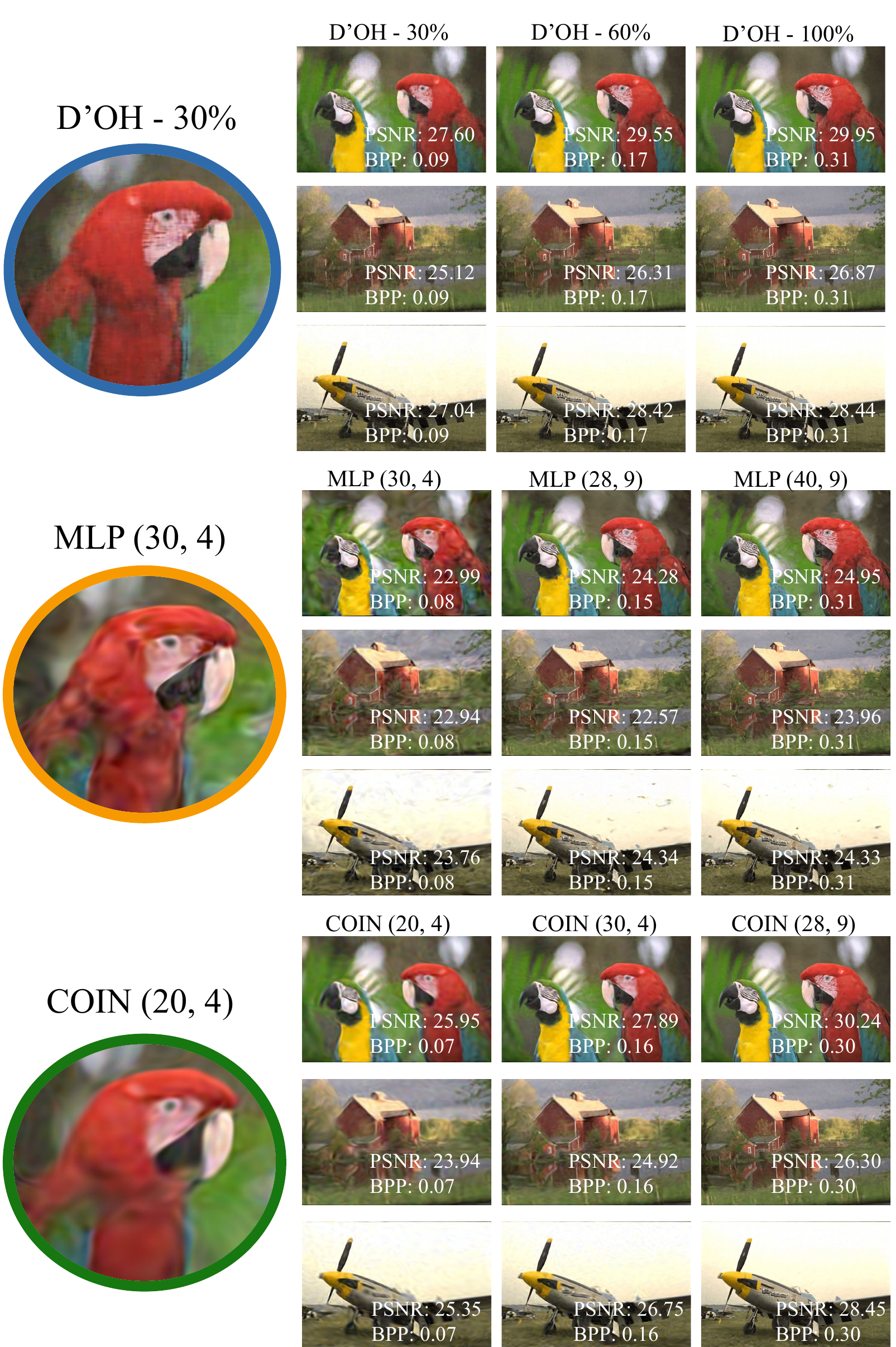}

    \captionof{figure}{Additional qualitative results on Kodak showing the comparison between 8-bit D'OH, 8-bit MLP, and COIN (a MLP quantized to 16-bits). Note that smaller COIN architectures are required to match the comparison bit-rates. D'OH is more robust to quantization than the MLP models. D'OH uses positional encoding, while the MLP models do not (see: Figure \ref{fig:pe_image} - PE is detrimental to low-rate MLP performance).}
    \label{fig:kodak_qual}
\end{minipage}

\clearpage
\subsection{Additional Qualitative Results - Occupancy Field}

\begin{figure}[h!]
    \centering
    \includegraphics[width=\linewidth]{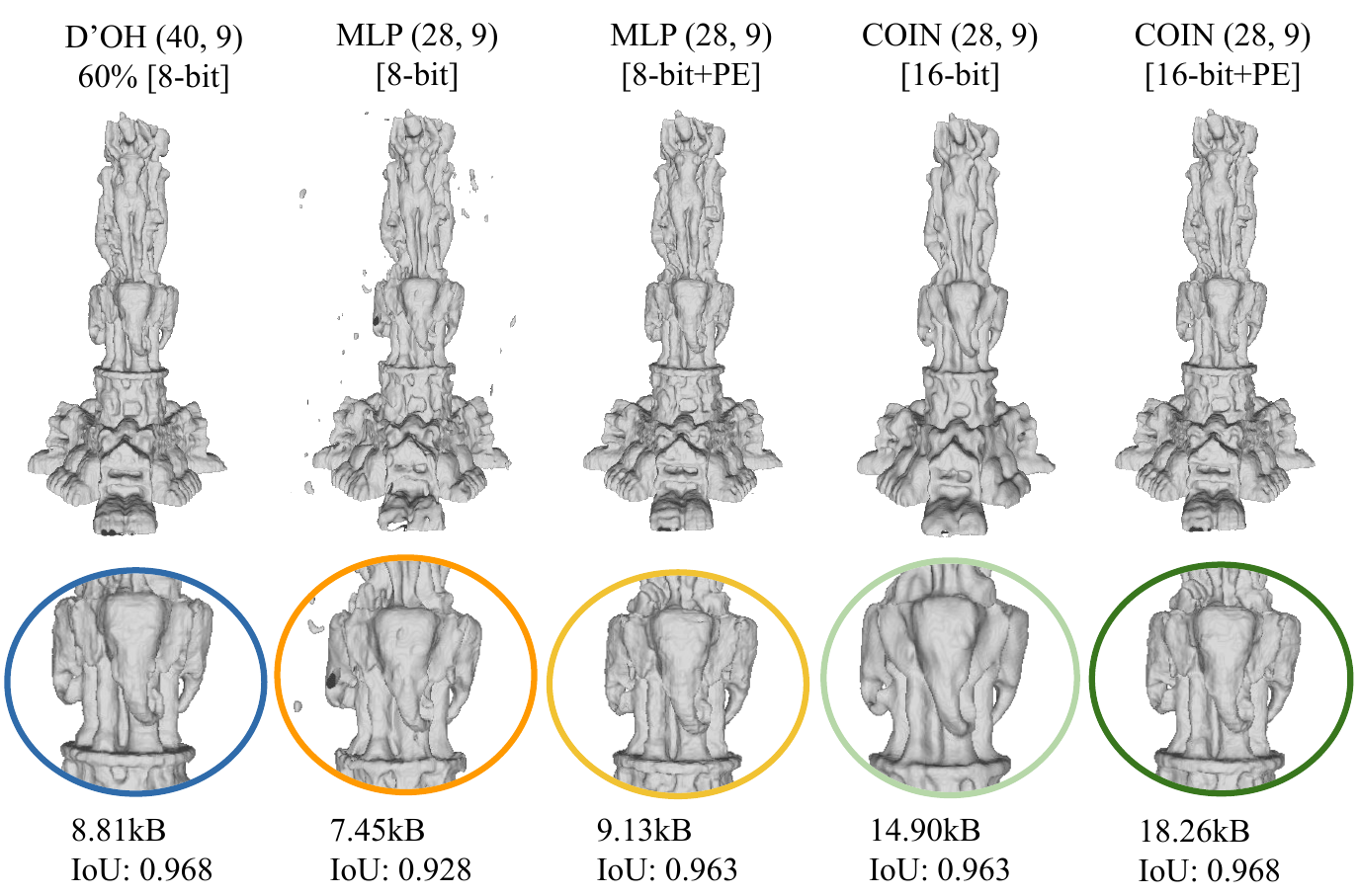}\\
    \includegraphics[width=\linewidth]{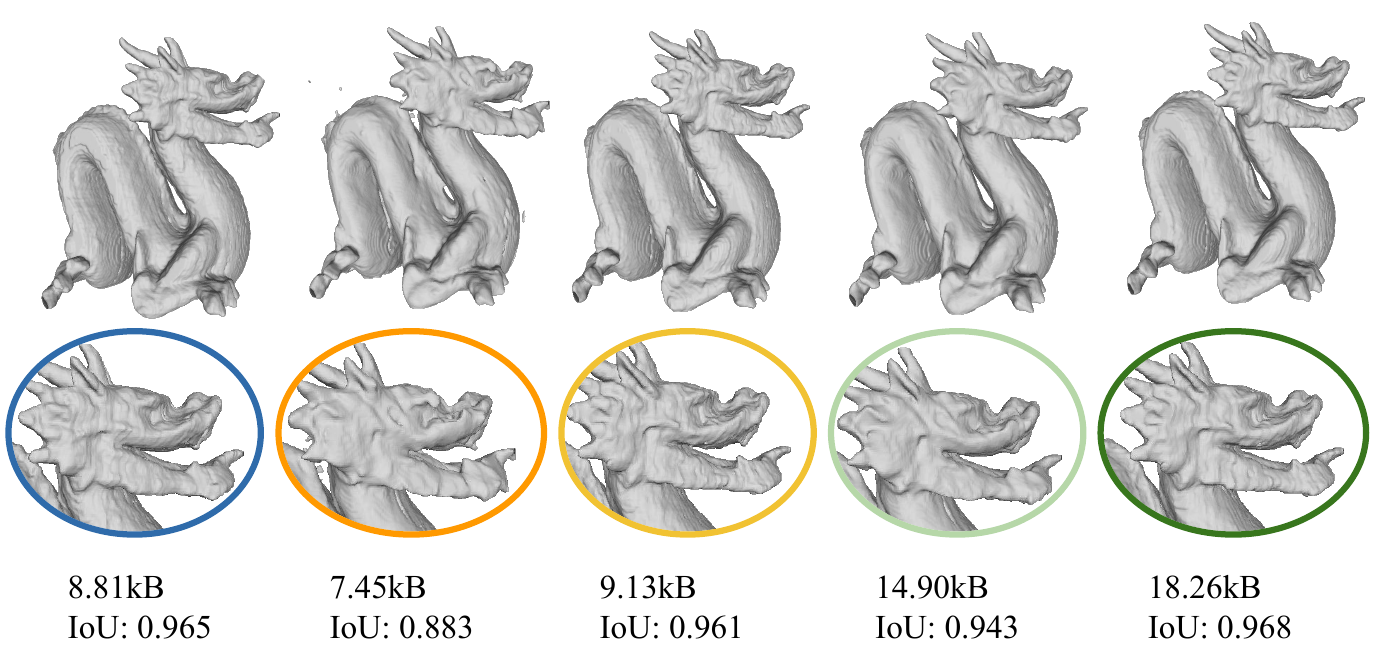}\\

    \caption{Binary Occupancy qualitative results on Thai Statue and Dragon. D'OH shows a large performance improvement over MLP models without positional encoding (which lose high frequency information), and shows a small rate-distortion improvement or equivalent performance to MLP and COIN models with higher memory footprints.}
    \label{fig:occupancy_qual}
\end{figure}

\clearpage 
\end{document}